\documentclass[pdflatex,sn-mathphys]{sn-jnl}

\usepackage{amsmath,amsfonts}
\usepackage{graphicx}
\usepackage{textcomp}
\usepackage{xcolor}

\usepackage{mathtools}
\DeclarePairedDelimiter{\norm}{\lVert}{\rVert}
\usepackage{breqn}

\usepackage{siunitx} 

\usepackage{numprint}

\usepackage{float}
\usepackage{dblfloatfix}
\usepackage{multirow}
\usepackage{etoolbox}

\graphicspath{ {./images/} }

\jyear{2022}%

\theoremstyle{thmstyleone}%
\theoremstyle{thmstyletwo}%

\theoremstyle{thmstylethree}%

\begin{document}

\title[Comparison and Analysis of Image-to-Image Generative Adversarial Networks]{Comparison and Analysis of Image-to-Image Generative Adversarial Networks: A Survey}


\author{\fnm{Sagar} \sur{Saxena}}\email{ssaxena1@umd.edu}

\author{\fnm{Mohammad Nayeem} \sur{Teli}}\email{nayeem@cs.umd.edu}

\affil{\orgdiv{Computer Science Department}, \orgname{University of Maryland}, \orgaddress{\city{College Park}, \state{Maryland}}}


\abstract{Generative Adversarial Networks (GANs) have recently introduced effective methods of performing Image-to-Image translations. These models can be applied and generalized to a variety of domains in Image-to-Image translation without changing any parameters. In this paper, we survey and analyze eight Image-to-Image Generative Adversarial Networks: Pix2Pix, CycleGAN, CoGAN, StarGAN, MUNIT, StarGAN2, DA-GAN, and Self Attention GAN.  Each of these models presented state-of-the-art results and introduced new techniques to build Image-to-Image GANs. In addition to a survey of the models, we also survey the 18 datasets they were trained on and the 9 metrics they were evaluated on. Finally, we present results of a controlled experiment for 6 of these models on a common set of metrics and datasets. The results were mixed and showed that, on certain datasets, tasks, and metrics, some models outperformed others. The last section of this paper discusses those results and establishes areas of future research. As researchers continue to innovate new Image-to-Image GANs, it is important to gain a good understanding of the existing methods, datasets, and metrics. This paper provides a comprehensive overview and discussion to help build this foundation.}




\maketitle

\section{Introduction}
\label{section:introduction}

The introduction of GANs by Goodfellow et al. \cite{gan} revolutionized how we can generate fake high-quality data with neural networks. Goodfellow et al. proposed the GAN that consists of two adversarial networks: a generator, which can capture the distribution of the data, and a discriminator, which can determine the probability that a sample is real or fake. GANs have progressed dramatically from the original models that were proposed by Goodfellow et al. Every aspect from the model architecture, loss functions, and inputs to the datasets we can use and metrics we can evaluate them on have been re-imagined and enhanced by researchers. 

Image-to-Image GANs have similarly revolutionized how we can generate hyper-realistic images. Instead of using a random distribution of data as input, these GANs use image inputs to generate other images. Image-to-Image GANs can be applied to numerous domains of Machine Learning and Computer Vision such as style transfer \cite{styletrans}\cite{stylegan}\cite{stylegan2}, colorizing \cite{pix2pix}, inpainting \cite{inpaint}, superresolution \cite{superres}\cite{superres2}, future state prediction \cite{future}, object transfiguration \cite{cyclegan}\cite{dagan}, photo editing and enhancement \cite{photoedit}\cite{photoedit2}\cite{cyclegan}, pose morphing \cite{dagan}, data augmentation \cite{dagan}, and many more. Pix2Pix \cite{pix2pix} was one of the first GANs that not only achieved Image-to-Image translation, but could also, in theory, be generalized to any dataset given sufficient training data and be able to model any of the aforementioned domains without having to alter the model. 

Previous survey papers on GANs \cite{gan_survey} and Image-to-Image synthesis \cite{img2img_survey} \cite{img_synth_and_editing_survey} include a wide array of models and a broad overview of different methods. This paper focuses on a small subset of models to present a more in-depth comparison and analysis. By focusing on this subset, this paper introduces new results on a common set of datasets and metrics for the models analyzed. While it would be ideal to extend this approach to all methods, it would be infeasible, as analyzing each model is a time-intensive process. 

This paper surveys eight Image-to-Image GANs that were proposed to be generalizable to any dataset: Pix2Pix, CycleGAN \cite{cyclegan}, CoGAN \cite{cogan}, StarGAN \cite{stargan},
MUNIT \cite{munit}, StarGAN2 \cite{stargan2}, DA-GAN \cite{dagan}, Img2Img Self-Attention GAN \cite{img2imgsa}. Each of these eight models introduced new advances in how Image-to-Image GANs are created and evaluated and set a new state-of-the-art in visual quality.

Each of these models has its own uniqueness in performing image-to-image translations. Some of these unique features can be summarized as:
\begin{itemize}
\item Pix2Pix was one of the first models that could be generalized to any dataset without altering its loss function. 
\item CycleGAN removed the limitation of needing paired data to train an Image-to-Image GAN. 
\item CoGAN introduced weight sharing to learn joint image distributions. 
\item StarGAN allowed Image-to-Image GANs to map between more than 2 classes with a single model and learn features common to all classes. 
\item MUNIT introduced multi-modal image translation to allow for one image to map to many styles. 
\item StarGAN2 expanded on MUNIT and StarGAN to perform multi-modal translation between more than two classes with a single model. 
\item DA-GAN was one of the first GANs to introduce the concept of using attention to generate higher quality images. 
\item Img2Img SA took advantage of the state-of-the-art SAGAN \cite{sagan} to introduce self-attention to image-to-image translation models.
\end{itemize}
As researchers continue to re-imagine how GANs can be further progressed to generate high quality data with the limitations of real-world data, it is important to understand what Image-to-Image GANs have accomplished in the past and how they have pushed the state-of-the-art. To provide the necessary background and analysis of Image-to-Image GANs, this paper:

\begin{itemize}
    \item surveys eight Image-to-Image GANs and compares the architectures and loss functions that were chosen,
    \item provides a brief discussion on the metrics that were implemented to evaluate these GANs and where they can be applied in the future, 
    \item provides an overview of the datasets that were used to train these models, and
    \item analyzes six Image-to-Image GANs and evaluates each model on a common set of datasets and metrics.
\end{itemize}

This paper is organized as follows: Section~\ref{section:models} covers each of the eight models in-depth and compares their architectures and loss functions with the other models. Section~\ref{section:metrics} examines each metric that was used to evaluate these Image-to-Image GANs. Section~\ref{section:datasets} describes all of the datasets that the eight models were trained on through a table of descriptions and figures that showcase samples from each dataset. Finally, in Section~\ref{section:analysis} we analyze six Image-to-Image GANs on four datasets chosen from Section~\ref{section:datasets} and two metrics chosen from Section~\ref{section:metrics}. 

\section{Models} 
\label{section:models}

\begin{figure}[!htb]
    \centering
    \includegraphics[width=0.75\linewidth]{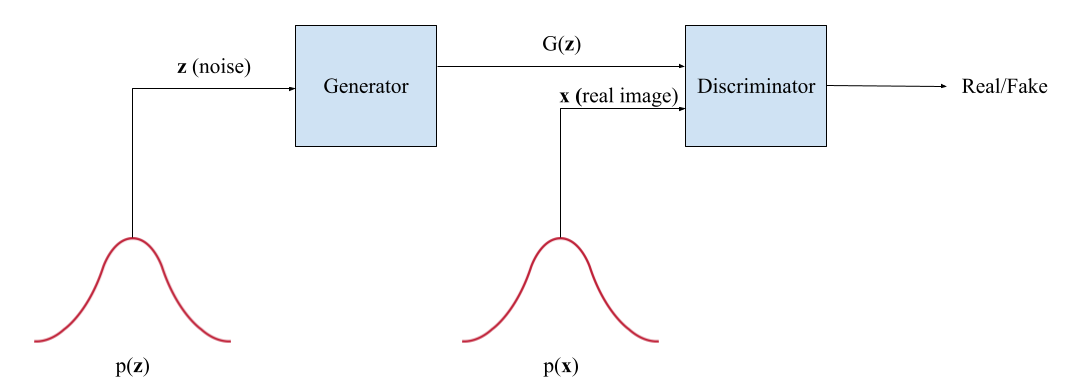}
    \caption{Basic Diagram of the Original GAN.}\label{fig:gan}
\end{figure}

The original GAN introduced by Goodfellow et al. \cite{gan} consisted of a generator, a network that captures the data distribution and generates realistic data, and a discriminator, a network that classifies whether the sample is generated or real. The generator's input is a latent vector, $z$, a randomly sampled noise vector, and the discriminator's input is the generated or real sample, $x$ (as depicted in Figure~\ref{fig:gan}). These two networks are trained to play a minimax game:

\begin{dmath}
arg\;\operatorname*{min}_{G}\;\operatorname*{max}_{D}
V(G, D) = \mathbb{E}_{x}[log D(x)] + \mathbb{E}_{z \sim p_z(z)}[log(1-D(G(z))]
\end{dmath}

Image-to-Image GANs are an extension of the original GAN that use image inputs to generate image outputs. The latent vector, $z$, is derived from, influenced by, or accompanied by the image inputs, which allows the generation of a corresponding image. This expands the ability of GANs to not only generate images from a single domain, but to also use the information from one domain to generate images in a second domain and achieve image-to-image translation. 

This section gives an overview of eight Image-to-Image GANs: Pix2Pix \cite{pix2pix}, CycleGAN\cite{cyclegan}, COGAN \cite{cogan}, MUNIT \cite{munit}, StarGAN\cite{stargan}, StarGAN2\cite{stargan2}, DA-GAN\cite{dagan}, Image-to-Image Self-Attention GAN \cite{img2imgsa}. For each model, we provide a brief history, a description of the architecture and objective functions, and its advantages and disadvantages compared to other models. 

\subsection{Pix2Pix}
Pix2Pix was one of the first Image-to-Image GANs that was able to generate high-quality images and serves as a foundation for many of the newer models that are discussed in this paper. Isola et al.~\cite{pix2pix} introduced the concept of automatic image-to-image translation: the task of translating one possible representation of a scene into another given sufficient training data. To achieve automatic image-to-image translation, Isola et al. proposed a novel concept of not only learning the mapping from an input image to an output image but also the loss function to train this model. This allows for a general approach to image-to-image translation that would traditionally require very different loss formulations.

Pix2Pix is comprised of a U-Net \cite{unet} based generator with skip connections and a PatchGAN based classifier. The U-Net architecture with skip connections prevents the bottleneck of low-level information which allows for the generation of higher quality images (as opposed to an encoder-decoder architecture). The discriminator penalizes structures at the scale of patches. Isola et al. demonstrated that the size of the patches can be significantly smaller than the size of the generated image, which allows the discriminator to more efficiently classify images as real or fake. In fact, their experiments showed that patches that were smaller than the size of the full image achieved higher accuracy in discriminating between real and fake images and produced higher quality images. 

The objective function for the conditional GAN mixes the traditional L1 loss function ($L_{L1}(G)$ with an adversarial loss ($L_{cGAN}(G,D)$) which is used to generate images indistinguishable from real images as shown below 
\begin{dmath}
L_{cGAN} (G, D) = \mathbb{E}_{x,y}[log D(x, y)] + \mathbb{E}_{x,z}[log(1-D(x,G(x,z))]
\end{dmath}
\begin{dmath}
L_{L1} (G) = \mathbb{E}_{x,y,z}[\norm{y-G(x,z)}_{1}]
\end{dmath}
\begin{dmath}
G^* = arg\;\operatorname*{min}_{G}\;\operatorname*{max}_{D} L_{cGAN}(G,D)+\lambda L_{L1}(G)
\end{dmath}
where, $G$ is the generator, $D$ the discriminator, $x$ is the input image, $y$ is the paired output image, $z$ is the Gaussian noise, and $\lambda$ is the relative importance of the $L1$ loss.

This model was able to achieve high-quality results with small datasets, but it required supervision at the level of paired images. The use of noise was also largely ignored by the model, and the researchers implemented dropout to simulate noise instead of an input noise vector. This resulted in low stochasticity and hindered how effective Pix2Pix would be outside of the scope of the training data.

\subsection{CycleGAN}
\label{subsection:cyclegan}
CycleGAN is an Image-To-Image GAN built on top of Pix2Pix that aimed to solve the problem of an absence of paired data when training an automatic image-to-image translation model. Pix2Pix focused on individual images and attempts to create a mapping that creates a generated image that is indistinguishable from the real image. CycleGAN, however, attempts to learn a mapping that creates a distribution of generated images that is indistinguishable from the distribution of real images. To do this CycleGAN finds special characteristics of one image collection and translates these characteristics to another image collection. 

CycleGAN is comprised of the same 70 x 70
PatchGAN discriminator from Pix2Pix \cite{pix2pix} and a generator adopted from previous work that had shown high-quality results with neural style transfer and super-resolution \cite{neustyle}. The generator was made with 6 to 9 residual blocks and two fractionally strided convolutions with stride 1/2. To exploit the set level characteristics of each collection, researchers implemented an adversarial loss, and a cycle consistent loss which consists of two parts: the forward consistency loss defined as $x \rightarrow G_{X \rightarrow Y}(x) \rightarrow F_{Y \rightarrow X} (G(x)) \approx x$ and the backward consistency loss defined as $y \rightarrow F(y) \rightarrow G (F(y)) \approx y$ where $F$ and $G$ are the generators for the $X$ and $Y$ domains respectively. As with Pix2Pix, both the mapping and the loss functions are trained. 
\begin{dmath}
L_{GAN}(G,D_B,X,Y) = \mathbb{E}_{y \sim p_{data}(y)}[log D_Y(y)] + \mathbb{E}_{x \sim p_{data}(x)}[log(1 - D_Y(G(x))]
\end{dmath}
\begin{dmath}
L_{cyc}(G,F) = \mathbb{E}_{x \sim p_{data}(x)}[\norm{F(G(x)) - x}_{1}] + \mathbb{E}_{y \sim p_{data}(y)}[\norm{G(F(y)) - y}_{1}] 
\end{dmath}
\begin{dmath}
L(G,F,D_X,D_Y)= L_{GAN}(G,D_Y,X,Y) + L_{GAN}(F,D_X,Y,X) + \lambda L_{cyc}(G,F)
\end{dmath}
\begin{dmath}
G*,F* = arg\;\operatorname*{min}_{G,F}\;\operatorname*{max}_{D_X,D_Y} L(G,F,D_X,D_Y)
\end{dmath}

In the equations above, $L_{GAN}$ is the adversarial loss which is very similar to Equation 1 from Pix2Pix with the key difference being the lack of a conditional GAN, and therefore the discriminator only takes in the generated or real output image as the input (and not the input image). $L_{cyc}$ defines the cycle consistency loss and sums forward consistency loss with backward consistency loss. The full objective function uses $\lambda$ to determine the relative importance of the cycle consistency loss and aims to solve G* and F*. 

Interestingly, CycleGAN used neither a U-Net based architecture opting for an autoencoder instead, nor skip connections, nor a conditional GAN (all of which were features of Pix2Pix). Overall, this model was able to achieve high-quality results with small datasets and only required supervision at the level of sets. However, Zhu et al. showed that in the cases with paired data, Pix2Pix outperformed CycleGAN. Furthermore, this model did not address the issue of stochasticity, which brings its effectiveness into question outside the scope of the training data. Nevertheless, the novel implementation of a cycle-consistent loss with Image-To-Image GANs is a feature that revolutionized future models. 

\subsection{CoGAN}
\label{subsection:cogan}
CoGAN is an image generation GAN that was developed and released in parallel to Pix2Pix that aimed to learn a joint distribution of multi-domain images. Similar to CycleGAN, CoGAN aims to generate a distribution of of images that is indistinguishable from the distribution of real images. CoGAN also uses an architecture of two generators and two discriminators, one for each domain. Rather than taking advantage of a cycle consistency loss (as in CycleGAN) to enforce that pairs of images are correlated, CoGAN introduces weight sharing to generate pairs of images. This allows a single noise vector to generate a pair of images from the two distributions. Notably, this model was not originally created for image-to-image translation, but Liu et al. extended their approach to image-to-image translation by finding the closest noise vector that generates an image closest to the input image and its translated pair.

CoGAN's generators, $G_1$ and $G_2$, comprise two identical deep fully convolutional networks, fractionally stride lengths, batch normalization, and parameterized rectified linear unit processing. The discriminators $F_1, F_2$ were a variant of the LeNet model \cite{lenet}. For both the discriminators and generators, Liu et al. found optimal results by enforcing wieght sharing for all layers except the last layer. The loss of the GAN consists of a combination of two adversarial terms:

\begin{dmath}
L_{adv}(F_1,F_2,G_1,G_2) = \mathbb{E}_{x_1 \sim p_{x_1}}[-log F_{1}(x_1)] + \mathbb{E}_{z \sim p_{z}}[-log(1 - F_1(G_{1}(z)))] + \mathbb{E}_{x_1 \sim p_{x_2}}[-log F_{2}(x_2)] + \mathbb{E}_{z \sim p_{z}}[-log(1 - F_2(G_{2}(z)))]
\end{dmath}

\begin{dmath}
G_{1}*, G_{2}* = arg\;\operatorname*{max}_{G_1,G_2}\;\operatorname*{min}_{F_1,F_2} L(F_1,F_2,G_1,G_2)
\end{dmath}

In the equations above, $x_1$ and $x_2$ are images sampled from their respective domains, $z$ is a latent vector sampled from the normal distribution $p_z$, $L_{adv}$ is the combination of two adversarial losses, which is similar to Equations 4 and 6 from CycleGAN. The full objective function aims to solve $G_{1}*$ and $G_{2}*$.

Overall, this model was able to achieve high quality results compared to other conditional GANs. The performance of this model on image-to-image translation tasks, however, was not rigorously tested and compared to other image-to-image translation models. The visual results provided by Liu et al. indicated that CoGAN is capable of image-to-image translation, but the authors stated that a more robust solution would need to be developed in future work. The contribution of weight sharing to learn joint multi-domain distributions is a feature that has been integrated into many Image-To-Image GAN's that have been released after CoGAN. 

\subsection{StarGAN}
StarGAN was introduced by Choi et al., who noted that prior models such as CycleGAN had achieved significant success with mapping one image domain to another, but these methods cannot be scaled beyond two domains. Choi et al. highlight how many datasets have various attributes that can be used for many domains in image-to-image generation. Previous techniques have focused on creating a GAN for each set of two domains, but not only are these techniques inefficient, they also fail to utilize the entire dataset and capitalize on common global elements. To solve this issue, StarGAN proposes a single generator method that jointly trains a mask vector that contains information about the domain. 

StarGAN is built on top of CycleGAN with two of the key differences being the loss function and the use of a conditional GAN. The objective function is comprised of an improved adversarial loss, a domain classification loss, and a reconstruction loss. The domain classification loss is included with an auxiliary classifier that ensures that an image is correctly translated from one domain to the other. The reconstruction loss is a cycle consistency loss that attempts to reconstruct an output image back to its original domain.

\begin{dmath}
L_{adv} = \mathbb{E}_{x}[D_{src}(x)] - \mathbb{E}_{x,c}[D_{src}(G(x,c))] - \lambda_{gp} \mathbb{E}_{\hat{x}}[(\norm{\triangledown_{\hat{x}}D_{src}({\hat{x}})}_{2} - 1)^2] 
\end{dmath}
\begin{dmath}
L_{cls}^r = \mathbb{E}_{x,c'}[- log D_{cls}(c'|x)]
\end{dmath}
\begin{dmath}
L_{cls}^f = \mathbb{E}_{x,c}[- log D_{cls}(c|G(x,c))]
\end{dmath}
\begin{dmath}
L_{rec} = \mathbb{E}_{x,c,c'}[\norm{x - G(G(x,c), c')}_{1}]
\end{dmath}
\begin{dmath}
L_D = - L_{adv} + \lambda_{cls} L_{cls}^r
\end{dmath}
\begin{dmath}
L_G = L_{adv} + \lambda_{cls} L_{cls}^f + \lambda_{rec} L_{rec}
\end{dmath}

where G is the generator that uses both an input image $x$ and a class vector $c$, $D_{src}$ classifies the source of an image and $D_{cls}$ classifies the class of an image. $L_{adv}$ is an improved version of the adversarial loss from Equation 4 used in CycleGAN called the Wasserstein GAN objective with gradient penalty \cite{wganobj}. The domain classification loss for real images is represented with $L_{cls}^r$ while $L_{cls}^f$ represents the domain classification loss for fake images where $c'$ is the original class vector and $c$ is the target class vector. $L_{rec}$ defines the reconstruction loss which ensures cycle consistency. The full objective function uses $\lambda_{cls}$ and $\lambda_{rec}$ for the relative importance of the reconstruction and domain classification losses. 

By being able to capitalize on the entire dataset and learn the global characteristics of all domains, StarGAN was able to outperform existing models and generate higher quality images. Most notably, StarGAN was able to translate images between multiple domains using a single generator network as opposed to previous networks which required at least two generators for each combination of domains. Finally, not only can StarGAN take advantage of a single dataset consisting of multiple domains, but it can also integrate multiple datasets that consist of different labeled domains. For example, one dataset of faces may consist of labels on hair color while another may consist of labels on eye color. StarGAN can train a model with both of these datasets due to its approach of using a masked vector that contains information about the domain. This feature allows StarGAN not only to translate between multiple domains, but also to generate higher quality images from data that do not need to be labeled. 

StarGAN allows for automatic image-to-image translation models with set-level semi-supervision to train more efficiently and achieve a higher level of accuracy than CycleGAN. It also seems to successfully integrate the concept of weight sharing for image-to-image translation by proposing a single generator that can capture multiple image distributions. But, as opposed to Pix2Pix and CycleGAN, this model does seem to require dataset-level supervision (i.e. shared global characteristics) to allow for it to achieve its full potential.

\subsection{MUNIT}
\label{section:munit}
MUNIT is an Image-to-Image GAN introduced by Huang et al. who noted that many image-to-image translation tasks are inherently multi-modal and an image in one domain can be represented by multiple images in a second domain. Huang et al. highlighted that prior approaches assume the mapping from one domain to another to be one-to-one and in doing so fail to produce diverse outputs. To solve this issue, MUNIT proposes that an image can be decomposed into a content code that is universal for all domains and a style code that is specific to each domain. 

MUNIT consists of an encoder $E_i$, a decoder $G_i$, and a discriminator $D_i$ for each image domain. Each encoder is broken into two separate components - one for generating the content code $E^c_i(x_i)$ and one for generating the style code $E^s_i(x_i)$. The content encoder consists of strided convolutional layers and residual blocks with each convolutional layer followed by instance normalization. The style encoder consists of strided convolutional layers, global average pooling, and fully connected layers. The decoder consists of a multilayer perceptron and adaptive instance normalization to decode the style code, residual blocks to decode both the content and style codes, and convolutional layers for upsampling. The loss of MUNIT consists of a bidirectional reconstruction loss, composed of an image reconstruction loss and a latent reconstruction loss, and an adversarial loss:

\begin{dmath}
L^{x_1}_{recon} = \mathbb{E}_{x_1 \sim p(x_1)}[\norm{G_1(E^c_1(x_1), E^2_1(x_1)) - x_1}_1]
\end{dmath}
\begin{dmath}
L^{x_2}_{recon} = \mathbb{E}_{x_2 \sim p(x_2)}[\norm{G_2(E^c_2(x_2), E^2_2(x_2)) - x_2}_1]
\end{dmath}
\begin{dmath}
L^{c_1}_{recon} = \mathbb{E}_{c_1 \sim p(c_1), s_2 \sim p(s_2)}[\norm{E^c_2(G_2(c_1, s_2)) - c_1}_1]
\end{dmath}
\begin{dmath}
L^{c_2}_{recon} = \mathbb{E}_{c_2 \sim p(c_2), s_1 \sim p(s_1)}[\norm{E^c_1(G_1(c_2, s_1)) - c_2}_1]
\end{dmath}
\begin{dmath}
L^{s_1}_{recon} = \mathbb{E}_{c_2 \sim p(c_2), s_1 \sim p(s_1)}[\norm{E^s_1(G_1(c_2, s_1)) - s_1}_1]
\end{dmath}
\begin{dmath}
L^{s_2}_{recon} = \mathbb{E}_{c_1 \sim p(c_1), s_2 \sim p(s_2)}[\norm{E^s_2(G_2(c_1, s_2)) - s_2}_1]
\end{dmath}
\begin{dmath}
L^{x_1}_{GAN} = \mathbb{E}_{c_2 \sim p(c_2), s_1 \sim p(s_1)}[log(1-D_1(G_2(c_2, s_1)))] + \mathbb{E}_{x_1 \sim p(x_1)}[log(D_1(x_1)))]
\end{dmath}
\begin{dmath}
L^{x_2}_{GAN} = \mathbb{E}_{c_1 \sim p(c_1), s_2 \sim p(s_2)}[log(1-D_2(G_2(c_1, s_2)))] + \mathbb{E}_{x_2 \sim p(x_2)}[log(D_2(x_2)))]
\end{dmath}
\begin{dmath}
arg\;\operatorname*{min}_{E_1,E_2,G_1,G_2}\;\operatorname*{max}_{D_1,D_2} L(E_1,E_2,G_1,G_2,D_1,D_2) = L^{x_1}_{GAN} + L^{x_2}_{GAN} + \lambda_x(L^{x_1}_{recon}+L^{x_2}_{recon}) + \lambda_c(L^{c_1}_{recon}+L^{c_2}_{recon}) + \lambda_s(L^{s_1}_{recon}+L^{s_2}_{recon})
\end{dmath}

In the equations above, $x_i$, $c_i$, and $s_i$ are the images, content codes, and style codes for domain $i$. $L^{x_i}_{GAN}$ is the adversarial loss for domain $i$. The three reconstruction terms for image, content code, and style code for each domain introduce cycle consistency to ensure that images and latent codes can be mapped back to their respective domains. The full objective function uses $\lambda_x$, $\lambda_c$, and $\lambda_s$ as weights that control the relative importance of the three reconstruction terms.

Overall, this model was able to achieve superior results to other image-to-image GANs such as CycleGAN and UNIT\cite{unit} and comparable results to GANs that required paired training data such as BicycleGAN\cite{bicyclegan}. The contribution of presenting a framework for unsupervised image-to-image translation that divides the latent space into content and style is a feature that reappears in future works.

\subsection{StarGAN2}

StarGAN2 iterates on StarGAN and arrives at the same conclusion that MUNIT does with the original StarGAN model; like other models, StarGAN did not capture the multi-modal nature of the underlying data and instead learned a deterministic one-to-one relationship. Prior approaches like MUNIT had solved this problem by decomposing the latent space into style and content codes, but they do not address the problems that StarGAN solved: these GANs do not scale well past two domains as it requires training multiple GANs and fail to utilize the entire dataset that may be available. To solve this, Choi et al. introduce StarGAN2, a framework that uses a single generator and discriminator with multiple branches to map between multiple domains and produce a diverse set of images. 

StarGAN2 consists of four modules: a generator $G$, a mapping network $F$, a style encoder $E$, and a discriminator 
$D$. The generator maps an input image $x$ from the origin domain and a style code $s$ from the target domain using AdaIN to an output image in the second domain. The mapping network maps a latent code $z$ to a domain-specific style code $\tilde{s}$. The style encoder maps an input image $x$ to it's corresponding style $s$. Finally, the discriminator maps an input image $x$ to a domain-specific real or fake binary value. All modules except the generator consist of multiple output branches for each domain. The objective function comprises an adversarial loss, style reconstruction loss, style diversification loss, and cycle consistency loss:
\begin{dmath}
L_{adv} = \mathbb{E}_{x,y}[log(D_x)] + \mathbb{E}_{x,\tilde{y},z}[log(1 - D_{\tilde{y}}(G(x, \tilde{s})))]
\end{dmath}
\begin{dmath}
L_{sty} = \mathbb{E}_{x,\tilde{y},z}[\norm{\tilde{s} - E_{\tilde{y}}(G(x, \tilde{s}))}_1]
\end{dmath}
\begin{dmath}
L_{ds} = \mathbb{E}_{x,\tilde{y},z_1,z_2}[\norm{G(x, \tilde{s_1}) - G(x, \tilde{s_2})}_1]
\end{dmath}
\begin{dmath}
L_{cyc} = \mathbb{E}_{x,y,\tilde{y},z_1}[\norm{x - G(G(x, \tilde{s}), \hat{s})}_1]
\end{dmath}
\begin{dmath}
arg\;\operatorname*{min}_{G,F,E}\;\operatorname*{max}_{D}\; L_{adv} + \lambda_{sty}L_{sty} - \lambda_{ds}L_{ds} + \lambda_{cyc}L_{cyc}
\end{dmath}

The adversarial loss uses a randomly sampled target domain $\tilde{y}$ and a randomly sampled latent code $z$ to calculate $\tilde{s} = F_{\tilde{y}}(z)$ and generate an image in the target domain. The style consistency loss introduces a reconstruction loss for the style codes. The style diversification loss enforces that images of two separate styles should be distinct. The cycle consistency loss enforces that an image mapped to a random target domain can be mapped back to the original image given its style code. 

Overall, this model was able to improve the baseline performance of StarGAN and present superior results to other image-to-image GANs such as MUNIT. By combining and advancing the approaches presented in StarGAN and MUNIT, researchers were able to both capture the multi-modal nature of the underlying data and create a model that can scale past two domains.

\subsection{Deep Attention (DA) GAN }
DA-GAN is an Image-to-Image GAN introduced by Ma et al., who noted that prior models such as CycleGAN favored set-level correspondences over instance-level correspondences. This preference impedes finding meaningful instance-level correspondences and can result in false positives and lead to mode collapse. It also limits the scope of a model outside of its training data. To exploit both set-level and instance-level correspondences, DA-GAN decomposes the task of translating samples from two independent sets into translating instances in a highly structured latent space via an attention mechanism. 

DA-GAN is comprised of a deep attention encoder, generator, and two discriminators. The Deep Attention Encoder learns a mapping function F and uses a loss of
\begin{dmath}
\mathbb{E}_{X \sim P_{data}(X)}[d(Y, DAE(X))]
\end{dmath}
where Y is the label of image X, and $d$ is some similarity metric in the data space. According to Ma et al. the choice for $d$ is flexible, as there are many options such as a VGG classifier. The instance-level representations from the deep attention encoder are concatenated and inputted to the generator, which consists of several residual blocks and a series of up-sampling layers. The discriminator consists of a series of down-sampling blocks and a fully connected layer to produce the decision score. One discriminator is used for the source domain and another is used for the target domain. The loss of the GAN consists of a consistency term, a symmetry term, and a multi-adversarial term:
\begin{dmath}
L_{cst} = \mathbb{E}_{s \sim P_{data}(s)}d(DAE(s), DAE(F(s))
\end{dmath}
\begin{dmath}
L_{sym} = \mathbb{E}_{t \sim P_{data}(t)}d(DAE(t), DAE(F(t))
\end{dmath}
\begin{dmath}
L_{GAN}^s = \mathbb{E}_{t \sim P_{data}(t)}[log D_1(t)] +  E_{t \sim P_{data}(s)}[log(1 - D_1(F(s)))]
\end{dmath}
\begin{dmath}
L_{GAN}^t = \mathbb{E}_{t \sim P_{data}(t)}[log D_2(t)] +  E_{t \sim P_{data}(t)}[log(1 - D_2(F(t)))]
\end{dmath}
\begin{dmath}
L(DAE,G,D_1,D_2)=L_{GAN}^s(DAE,G,D_1,S,T) + L_{GAN}^t(DAE,G,D_2,T) + \alpha L_{cst}(DAE,G,S) + \beta L_{sys}(DAE, G, T)
\end{dmath}
\begin{dmath}
F* = \operatorname*{argmin}_{F}\;\operatorname*{max}_{D_1,D_2} L(F,D_1,D_2)
\end{dmath}

where $DAE$ is the deep attention encoder that maps an input image into a latent code and $F$ is a function that maps a sample from domain $S$ to domain $T$. $F$ uses $DAE$ to first encode an image from domain $S$ to a latent code and then $G$ to decode that latent code into an image in domain $T$.  $L_{cst}$ represents the consistency loss and $L_{sym}$ is the symmetry loss that ensures that F can map samples from T to itself.

$L_{GAN}^s$ is the adversarial loss that is most similar to Equation 4 from CycleGAN that ensure that translated samples are indistinguishable from real samples while $L_{GAN}^t$ is the adversarial loss that helps prevent mode collapse by ensuring that when mapping samples from T to themselves, F generates closely located modes. The full objective function uses $\alpha$ and $\beta$ for the relative importance of the consistency loss and symmetry loss and aims to solve F*. 

By being able to capture both set-level and instance-level correspondences, Ma et al. showed that DA-GAN achieved a higher visual quality than prior methods such as CycleGAN. DA-GAN was also able to be employed in domains of object transfiguration and pose-morphing, two domains that are difficult to address with CycleGAN which is able to perform better at style transfer. This shows that DA-GAN is also able to capture higher stochasticity. Although it is outside the scope of this paper, it is interesting to note that DA-GAN is the only model in this paper that has been tested outside of Image-to-Image translation and showed superior results in Text-to-Image synthesis as well. It is important to note that, to the best of our knowledge, Ma et al. never open-sourced the code associated with their research, which reduces how accessible it is to adopt and compare with other methods listed in this paper. Nevertheless, the implementation of a deep attention mechanism seemed to push the state-of-the-art. 

\subsection{Img2Img SA}

The Img2Img SA network is an Image-to-Image GAN introduced by Kang et al. who noted that prior methods like CycleGAN and MUNIT had achieved good results by changing low level information, but fail to change high level information. Kang et al. highlights that these prior methods have been very successful in style transfer and many-to-many mapping between image domains where the latent space can be decomposed into style and content codes. Image-to-Image translation tasks, such as object transfiguration, that may require strong geometric change, are not modeled well by these style-based methods. Inspired by SAGAN which demonstrated that a self-attention module can help model long range, multi-level dependencies, Kang et al. proposed an unpaired Image-to-Image translation network with self-attention networks to change high-level information when mapping between two domains.

The Img2Img SA network is built on top of MUNIT and includes several self-attention blocks in both the generator and discriminators. In the generator, a self-attention layer is placed before the downsampling layer in the encoder and after the upsampling layer in the decoder. In the discriminator, the self-attention layer is placed before the downsampling layer. No other modifications are made to the model architecture and loss functions described in Section~\ref{section:munit}. The self-attention module is defined as
\begin{dmath}
\beta_{j,i} = \frac{exp(s_{ij})}{\sum_{i=1}^N exp(s_{ij})}
\end{dmath}

\begin{dmath}
o_j = \sum_{i=1}^N \beta_{j,i}h(x_i) 
\end{dmath}
where $s_{ij} = f(x_i)^T g(x_j)$, $f(x_i) =  W_f x_i$, $g(x_i) = W_g x_i$, $h(x_i) = W_h x_i$. In the equations above, $f$ and $g$ are the feature spaces used to calculate attention, $o = (o_1, o_2, ..., o_n)$ is the output of the attention layer, and all weight parameters $W$ are represented by 1x1 convolutions.

By introducing self-attention to the MUNIT model, this approach was able to outperform prior methods such as CycleGAN and MUNIT. By combining the decomposition of the latent space into style and content with the advances in self-attention for image generation, researchers were able to capture both low-level and high-level transformations between two domains and validate the use of self-attention for Image-to-Image GANs.  
\section{Metrics}
\label{section:metrics}

The following section provides a brief survey on the metrics used by the eight Image-to-Image models. These studies have shown that choosing only one metric to demonstrate the effectiveness of a model is usually insufficient. Each model surveyed used a combination of the metrics below to provide a more effective measure of its performance. Overall, Fr\'echet Inception Distance\cite{fid} and Precision and Recall \cite{precrec} seem to be two of the most accessible and effective methods for any Image-to-Image GAN, but all methods below can be leveraged to provide better quantitative and qualitative performance measures of GANs and a combination of multiple metrics should always be chosen that best represents the dataset or model. In general, metrics should be chosen that capture both the quality and stochasticity of the generated output.

\subsection{Amazon Mechanical Turk}
Amazon Mechanical Turk allows researchers to assign tasks for users (Turkers) to perform at scale. In the context of Image-to-Image generation, it can be used as an accessible method for qualitatively evaluating the performance of models. Through this method, Turkers are usually prompted to evaluate which images are fake and which images are real from the generated and real datasets. This method was employed by Pix2Pix, CycleGAN, and StarGAN. Although this metric is useful in determining how realistic generated images look to humans, it is a binary metric that only represents how likely the generated image would be perceived as real, rather than being a quantitative measure of image quality or stochasticity. This metric is best accompanied with a quantitative measure that is able to capture those features. 

\subsection{Domain Adaptation Score}
The domain adaptation score \cite{cogan} is applicable when a classifier can be trained to identify the class that a generated image belongs to. This approach was employed by CoGAN; CoGAN was trained on the MNIST\cite{mnist} and USPS\cite{usps} datasets jointly with a digit-classifier that was trained on MNIST. This digit classifier was then applied to the USPS dataset and the accuracy of this classifier was reported as the evaluation of CoGAN.

StarGAN used a similar approach to classify the facial expression that the model generated on the RAFD dataset. An expression classifier was trained and evaluated using a 90/10 split and StarGAN was fit on the training data to output faces with different expressions. The accuracy of the classifier on the generated expressions on the test set was then reported as the evaluation of StarGAN.

\subsection{FCN-Score}
The FCN-Score \cite{pix2pix} is applicable when a fully convolutional semantic classifier can be trained with a given dataset and the output segmentation map of that classifier can be compared for real and generated images \cite{proscons}. This method is best employed when there is paired data available and a segmentation map is being used to generate an image. The FCN score would then represent the accuracy of the classifier on the generated image compared to the input segmentation map.  Pix2Pix and CycleGAN used it for both per-pixel and per-class accuracy scores.

\subsection{Inception Score}
The Inception Score uses a pre-trained Inception V3 model, a network that can classify 1000 objects, to classify generated images \cite{noteincep} that was employed by DA-GAN. These classifications are summarized in the Inception Score which captures both the quality and diversity of images \cite{tragan}. The initial metric for the Inception Score was: 
\begin{dmath}
IS(G) = exp(\mathbb{E}_{\textbf{x} \sim p_g} D_{KL}(p(y|\textbf{x})||p(y)))
\end{dmath}
where $p(y|\textbf{x})$ is the conditional label distribution computed through the Inception V3 model, $p(y) = \int_{x} p(y|x)p_g(x)$ is the marginal class distribution, $D_{KL}$ is the Kullback-Leibler divergence, and $p_g$ is the distribution of the generated images \cite{tragan}\cite{noteincep}. This score was later improved to be:

\begin{dmath}
S(G) = \frac{1}{N} \sum_{i=1}^{N} D_{KL}(p(y|x^{i}) || \hat{p}(y))
\end{dmath}
where N is the number of sample images, $\hat{p}(y) = \sum_{i=1}^{N} p(y|x^{i})$ is the empirical marginal class distribution calculated over the entire generated dataset, and the exponentiation is removed.

The Inception Score was, however, shown to be sensitive to weights, has known issues when applied beyond the original ImageNet dataset, and doesn't account for models overfitting to a subset of the training data \cite{noteincep}. 

\subsection{Missing Modes}
Missing modes is a metric that was employed by DA-GAN that was used to represent the number of modes that the Image-to-Image GAN never generated. Similar to the FCN-Score, this method requires a classifier that has been trained on the dataset to determine the mode to which an image belongs. Ma et al. used this method in DA-GAN to supplement the Inception Score and provide a metric for measuring if a model was overfitting. 

\subsection{Conditional Inception Score}

The conditional inception score \cite{munit} is a modified inception score proposed by Huang et al. for multi-modal translation. This score is able to better represent the diversity of output images by recognizing that each input image can be mapped to multiple modes. The score is defined as 

\begin{dmath}
CIS = \mathbb{E}_{x_1 \sim p(x_1)}[\mathbb{E}_{x_{1\rightarrow 2} \sim p(x_{2 \rightarrow 1} \bar x_1)}[KL(p(y_2 \bar x_{1 \rightarrow 2}||p(y_2))]] 
\end{dmath}
where $p(y_2 \bar x_2)$ is a classifier that classifies an image $x_2$ in its mode $y_2$. 

This method, however, requires supervision at the level of the number of modes in a domain and the ground truth label for each image and its mode to train the classifier $p$. 

\subsection{Fr\'echet Inception Distance}
\label{subsection:fid}
Fr\'echet Inception Distance (FID) measures differences in the density of two distributions in the high-dimensional feature space of an InceptionV3 Classifier \cite{stylegan2}. Instead of solely focusing on the generated images like the Inception Score, the FID compares the distribution of real images and generated images \cite{fid}. The metric for the FID is
\begin{dmath}
d^{2}((m, C), (m_w, C_w)) = \norm{m - m_w}^2_{2} + Tr(C+C_w - 2(CC_w)^{1/2})
\end{dmath}
where $(m, C)$ represents the mean and covariance matrix of the Gaussian distribution that represents the inception scores of the generated data, $(m_w, C_w)$ represents the mean and covariance matrix of the Gaussian distribution that represents the inception scores of the real data, and $\textit{Tr}$ represents the trace \cite{fid}. 

This metric improves the Inception Score by using the real dataset as well as the generated dataset which would help account for overfitting and the sensitivity to weights. Later studies did show that classifiers trained with ImageNet, such as InceptionV3, tend to favor textures over shapes and which does impede FID's ability to accurately represent the quality of an image \cite{imnettext}. This metric was used to evaluate StarGAN2. 

\subsection{Precision and Recall}
\label{subsection:prerec}
Precision and Recall \cite{precrec} was introduced to supplement evaluation metrics like Fr'echet Inception Distance, which provide only a single numerical value for defining the performance of a metric and do not provide more detailed information on how that performance can be improved. This metric introduced two new definitions of precision and recall in the context of GANs. Given a generated distribution Q and a reference distribution P, precision measures how much of Q can be generated by P and recall measures how much of P can be generated by Q. The formal definition given by Mehdi et al. is:
\begin{dmath}
P = \beta \mu + (1 - \beta) \nu_P
\end{dmath}
\begin{dmath}
Q = \alpha \mu + (1 - \alpha) \nu_Q
\end{dmath}
where $\alpha$ is the precision and $\beta$ is the recall given that there exist distributions $\nu_P$, $\nu_Q$, and $\mu$. $\mu$ is the true common component of $P_S$ and $Q_S$, $\nu_P$ is the part of P that is missed by Q, and $\nu_Q$ is the part of Q that is missed by P. 

Overall this method seems like a good supplement to other metrics as it not only demonstrates the quality of images being generated but also provides insight into where the model may be lacking (failing to generate images that are similar to the original dataset or failing to capture the full entropy of the original dataset). 

\subsection{LPIPS Distances}
The Learned Perceptual Image Patch Similarity (LPIPS) Distances metric was originally introduced as a measure of perceptual loss \cite{percepmetr}. LPIPS incorporates deep learning features from VGG, AlexNet, and SqueezeNet architectures to provide a metric that computes the distance between paired images.

It was used by MUNIT and StarGAN2 as a measure of the average pairwise distances between all generated outputs from the same source image:
\begin{dmath}
\frac{2}{N(N-1)} \sum_{i=1}^N \sum_{j=i+1}^N D_{LPIPS}[G(x, s_i), G(x, s_j)]
\end{dmath}
where $x$ is the source image and $s_i$ is a randomly sampled style vector. This metric is aggregated as a single value by averaging over all test images. Both MUNIT and StarGAN2 used LPIPS to represent the diversity of the generated images, with higher LPIPS scores corresponding to a greater diversity.

LPIPS is an effective measure, but it seems limited to models where there is either supervision at the level of pairs or many-to-many image translation.
\section{Datasets}
\label{section:datasets}

This section provides a concise overview in a tabular format of the datasets used by the eight Image-to-Image models described in Section~\ref{section:models}. Instead of presenting them in the following paragraphs as prose we explore them in the format of a table. This presentation gives a meaningful snapshot of the models and their datasets in a single view. To this effect, Table~\ref{tab:datadesc} lists a total of 18 datasets that these eight models used.  

In this table, the first column is a data set name, the second column provides a brief description of each dataset, including the type of data, the number of images, and how the dataset is annotated, and the third column lists the models that were trained on the specific datasets. The fourth column lists how that model is applied to the dataset. 

\begin{figure*}[hp]
\caption{Sample images from datasets from Table~\ref{tab:datadesc}. Six images were sampled from each dataset. For all datasets except ImageNet, UT Zappos50k, and Sketch, the six images constitute 3 pairs of images. The row for Cityscapes has each original image from a city followed by its semantic labels. The row for Facades has an image of a facade followed by an image of its labels. The row for Transient Attributes has an image taken during the night followed by one taken during the day. The Multispectral Pedestrian row has a color image followed by its paired thermal image. Finally, the Google Maps row showcases an image of a satellite map followed by its paired standard map image. With the exception of the image of the dog and cat from ImageNet, all images were randomly sampled.}
\label{fig:sampleori}
\centering
\includegraphics[width=1.0\textwidth]{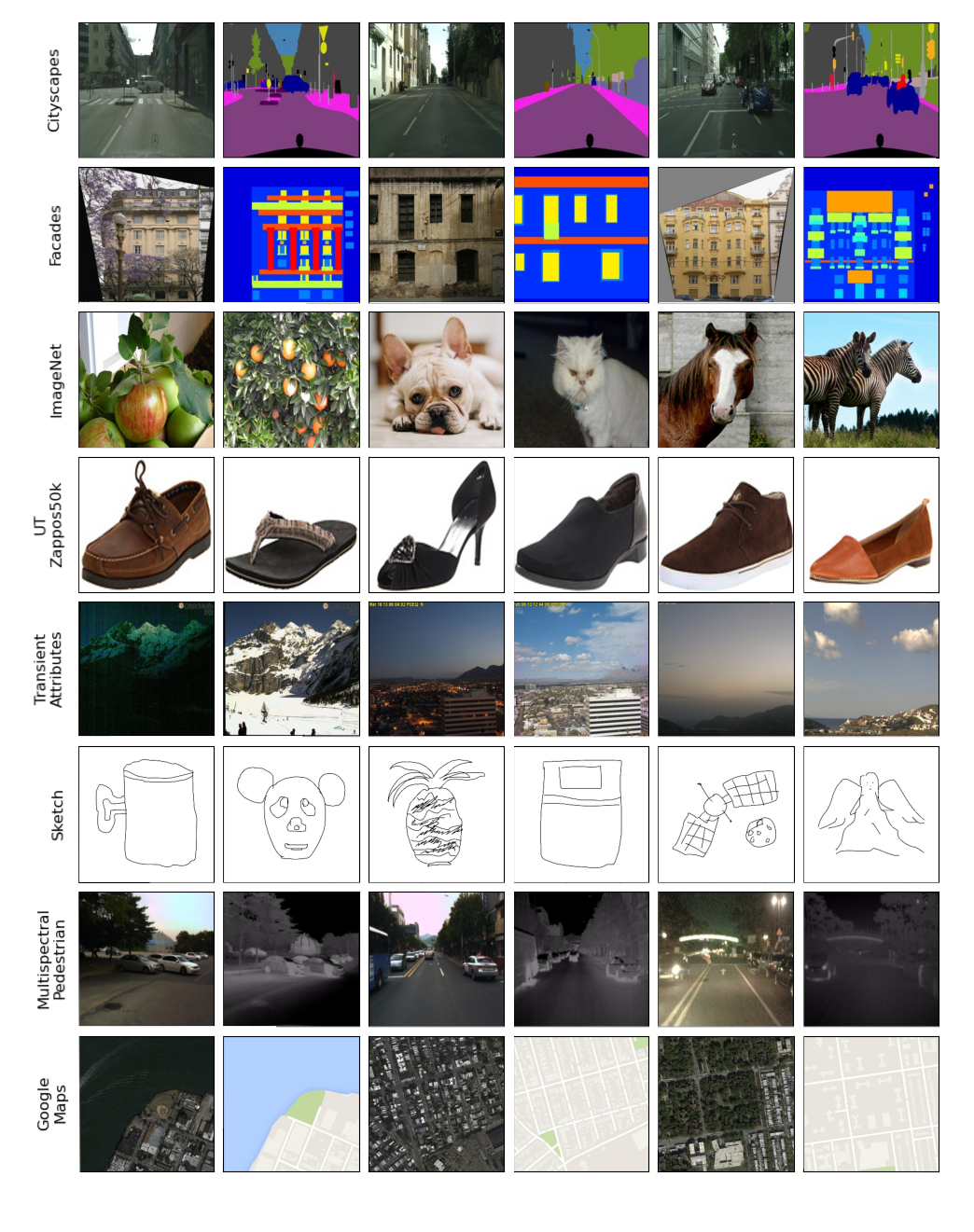}
\end{figure*}

\begin{figure*}[hp]
\caption{Sample images from datasets from Table~\ref{tab:datadesc} continued. Six images were sampled from each dataset. Images from Flickr were randomly sampled from the datasets on Flickr provided by CycleGAN. The images from NYU-Depth-V2 and Radboud Faces were taken directly from the samples hosted at their official sites. The images from SYNTHIA were taken from samples included in \cite{synthia}. All other images were randomly sampled from their respective datasets.}
\label{fig:sampleori2}
\centering
\includegraphics[width=1.0\textwidth]{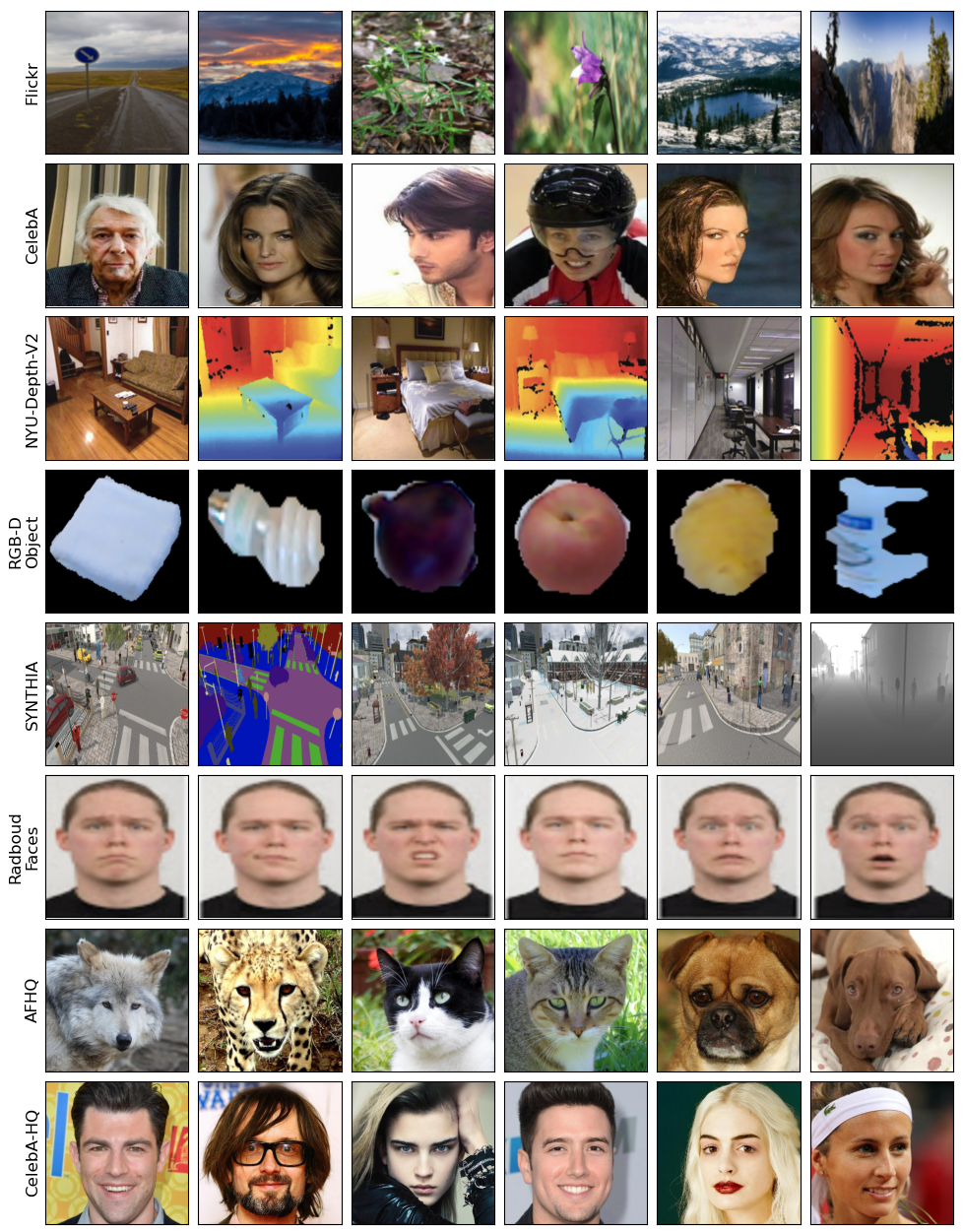}
\end{figure*}

\begin{figure*}[ht]
\caption{Sample images from datasets from Table~\ref{tab:datadesc} continued. Six images were randomly sampled from each dataset.}
\label{fig:sampleori3}
\centering
\includegraphics[width=1.0\textwidth]{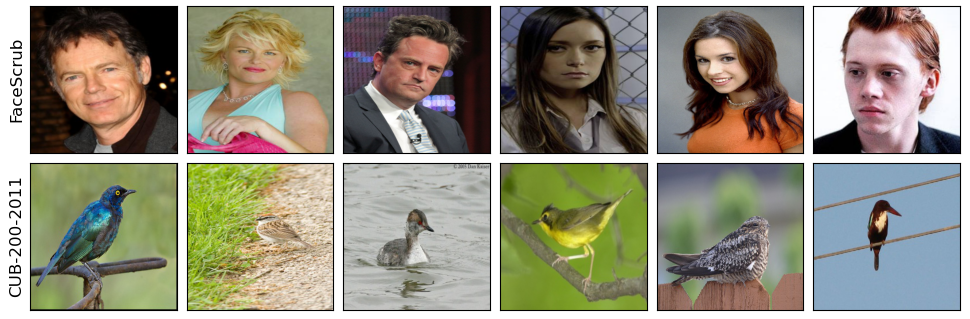}
\end{figure*}

Figures~\ref{fig:sampleori},~\ref{fig:sampleori2}, and~\ref{fig:sampleori3} show six representative samples from each dataset. This list is by no means exhaustive, but these datasets cover a fair range of diverse data to evaluate the performance of new Image-to-Image GANs and establish baselines using existing models. 

\section{Results}

\begin{table*}[htbp]
  \scriptsize
  \begin{center}
    \caption{Descriptions and Applications of Datasets}
    \label{tab:datadesc}
    \resizebox{\textwidth}{!}{%
    \begin{tabular}{p{2.5cm} p{6cm} p{1.6cm} p{4cm}}
      \textbf{Dataset} & \textbf{Description} & \textbf{Models} & \textbf{Applications}\\
      \hline
      {Cityscapes\cite{cityscapes}} & {25000 images of 50 cities with annotations for semantic segmentation across 30 classes.} & {Pix2Pix \newline CycleGAN \newline MUNIT} & {$Semantic\;Labels \leftrightarrow Photo$}\\
      \hline
      {CMP Facade\cite{facades}} & {606 images of facades from various cities that have been labeled with 12 classes.} & {Pix2Pix \newline CycleGAN \newline DA-GAN} & {$Architectural\;Labels \leftrightarrow Photo$}\\
      \hline
      \multirow[t]{2}{=}{ImageNet\cite{imgnet}} & {\multirow[t]{2}{=}{A large, diverse, and growing database of over 14 million images that aims to collect 1000 image urls and thumbnails for each of the 100000 synsets in WordNet \cite{wordnet}.}} & Pix2Pix & {$Black\;And\;White \rightarrow Color$} \\\cline{3-4}
      & & {CycleGAN} & {$Apple \leftrightarrow Orange$ \newline $Horse \leftrightarrow Zebra$ \newline $Dog \leftrightarrow Cat$ \newline}\\
      \hline
      {UT Zappos50k \cite{utzappos}} & {50025 images of shoes collected from Zappos.com that have 4 major categories - shoes, sandals, slippers, and boots - and 4 relative attributes - open, pointy, sporty, and comfort.} & {Pix2Pix \newline CycleGAN \newline MUNIT \newline Img2Img SA} & {$Edges \leftrightarrow Shoe$}\\
      \hline
      {Transient Attributes \cite{transient}} & {8571 images taken from 101 webcams that have been annotated with 40 attribute labels mainly for transient scene annotations.} & {Pix2Pix} & {$Day\rightarrow Night$}\\
      \hline
      {Sketch \cite{sketch}} & {20000 sketches of 250 everyday objects.} & {Pix2Pix} & {$Sketch \rightarrow Photo$}\\
      \hline
      {Multispectral Pedestrian \cite{multispec}} & {95000 color and thermal images of regular traffic scenes with bounding box annotations for pedestrians.} & {Pix2Pix} & {$Thermal \rightarrow Color$}\\
      \hline
      {Google Street View and Google Maps} & {Provide valuable data on maps and images of locations globally that can be scraped.} & {Pix2Pix} & {$Map \leftrightarrow Aerial$ \newline $Photo\;with\;Missing\;Pixels \rightarrow Inpainted\;Photo$}\\
      \hline
      \multirow[t]{3}{=}{Flickr} & {\multirow[t]{3}{=}{Website that hosts tens of billions of photos that can be scraped.}} & {CycleGAN} & {$Photograph \leftrightarrow Artist's\; Rendition$ \newline $Summer \leftrightarrow Winter$ \newline $Flower \leftrightarrow Enhanced\; Flower$} \\
      \cline{3-4}
      & & {DA-GAN} & {$Flower \leftrightarrow Enhanced\; Flower$}\\
      \hline
      \multirow[t]{2}{=}{CelebA \cite{celeba}} & {\multirow[t]{2}{=}{202599 face images of 10177 celebrities with annotations for 5 landmarks and 40 binary attributes.}} & {CoGAN \newline StarGAN} & {Map Between Different Binary Attributes.}\\
      \\\cline{3-4}
      & & Img2Img SA & {$Face \rightarrow Dog$ \newline $Face \rightarrow Cat$ \newline $Portrait \rightarrow Face$}\\
      \hline
      {NYU-Depth-V2\cite{nyu}} & {1449 pairs of aligned RGB and depth images of a variety of indoor scenes with dense multi-class labels and 407024 unlabeled RGB and Depth pairs.} & {CoGAN} & {Generate RGB and Depth Pairs.}\\
      \hline
      {RGB-D Object\cite{rgbd}} & {250000 images of 300 common household objects divided into 51 categories with category, instance, and pose labels.} & {CoGAN} & {Generate RGB and Depth Pairs.}\\
      \hline
      {SYNTHIA\cite{synthia}} & {200000 images from video streams and 20000 images from independent snapshots of a photo-realistic virtual city with pixel-level semantic segmentation for 13 classes.} & {MUNIT} & {$Cityscape \leftrightarrow SYNTHIA$}\\
      \hline
      {Radboud Faces \cite{rafd}} & {Face images of 67 models displaying 8 emotional expressions, each in 3 different gazes with 5 different camera angles.} & {StarGAN} & {Map Between Different Facial Expressions.}\\
      \hline
      {AFHQ\cite{stargan2}} & {15000 512x512 images of 3 domains - cats, dogs, and wildlife - with at least 8 breeds in each domain.} & {StarGAN2} & {Map between different animal domains}\\
      \hline
      {CelebA-HQ\cite{proggan}} & {High quality version of CelebA \cite{celeba} with 30000 1024x1024 images of faces.} & {StarGAN2} & {Map Between Different Binary Attributes}\\
      \hline
      {FaceScrub \cite{facescrub}} & {106863 face image urls of 530 people.} & {DA-GAN} & {$Human\;Face \rightarrow Animation$}\\
      \hline
      {CUB-200-2011 \cite{cub200}} & {11788 images of 200 different birds with annotations for 15 part locations, 312 binary attributes, and 1 bounding box per image.} & {DA-GAN} & {$Source \rightarrow Target\; Species$}\\
    \end{tabular}}
  \end{center}
  Table~\ref{tab:datadesc}: Descriptions and applications of the 18 publicly available datasets from Section~\ref{section:datasets}. For each dataset, a description of the dataset, the models that used that dataset, and the applications those models created with the specified dataset are provided above. Each model has been trained for applications that align with its cell. If there are multiple models listed but they are not separated, each model was trained for all accompanying applications (i.e. both Pix2Pix and CycleGAN were trained for a $Semantic\;Labels \leftrightarrow Photo$ application on the Cityscapes dataset).
\end{table*}

\label{section:analysis}
In this section, we present results on four datasets described in Section~\ref{section:datasets} by running multiple experiments with six models explained in Section~\ref{section:models}. These results are evaluated on two metrics (explained below) presented in Section~\ref{section:metrics}. The Methods subsection below details the setup of these experiments and the Analysis subsection details our results and findings.  

\subsection{Methods}
We chose to compare six models - Pix2Pix, CycleGAN, CoGAN, MUNIT, Img2Img SAGAN, and StarGAN2. Due to the absence of open-sourced code for DA-GAN and the fact that StarGAN is an earlier version of StarGAN2, those two models have been left out from our analysis. These models were trained on four datasets - UT Zappos50k, Transient Attributes, CUB-200-2011, and CelebA. We chose the CUB-200-2011 dataset as it was used by DA-GAN but not by the other four models which makes it a new dataset for all models. Given that Pix2Pix requires paired data, we also chose the UT Zappos50k and Transient Attributes datasets. Finally, we chose the CelebA dataset as it is one of the biggest datasets that offers a wide variety of attributes to train on. 

For each dataset, a 90-10 train-test split was made where the data was randomly sampled for the split. Since the Transient Attributes dataset only has 101 unique locations, these locations were split between the train and test datasets. More information on each of these datasets can be found in Section~\ref{section:datasets}. The results were reported across two metrics, Fr'echet Inception Distance and Precision and Recall. We chose these metrics as they are two of the most accessible and effective metrics available to analyze the performance of Image-to-Image GANs (see Sections~\ref{subsection:fid} and ~\ref{subsection:prerec} respectively). FID, precision, and recall were calculated with the output images as one image distribution and the original images as the second image distribution.  

\begin{table}[!h]
  \begin{center}
    \caption{Training Parameters, Approximate Time per Kimg, and Epochs}
    \label{tab:restime}
    \begin{tabular}{l S S r}
      \textbf{Model} & \textbf{Parameters} & \textbf{Time (s)/kimg} & \textbf{Epochs}\\
      \hline
      {Pix2Pix}    & {57.2M}   & {74}  & {75}\\
      {CycleGAN}   & {114.5M}  & {208} & {25}\\
      {CoGAN}      & {27.6M}   & {13}  & {n/a}\\
      {MUNIT}      & {46.6M}   & {118} & {50}\\
      {StarGAN2}   & {77.7M}   & {509} & {50}\\
      {Img2Img SA} & {48.7M}   & {525} & {25}\\
    \end{tabular}
  \end{center}
\end{table}

Each model was trained and evaluated on 4 NVIDIA GeForce GTX 1080 Ti GPUs.  The number of parameters for each model, the approximate time taken per kimg (1000 images) in seconds, and the number of epochs each model was trained for, are listed in Table~\ref{tab:restime}. The number of epochs that each model should be trained for was calculated by determining the epoch at which the model's FID had converged. All images were scaled to 256x256 for each dataset. An exception was made in the case of CoGAN, which, due to its fully convolutional model architecture, required 128x128 images.

CoGAN was the lightest model, taking only 13 seconds to process 1000 images. But, this was due to the fact that with image sizes limited to 128x128, the model could use a batch size of 128. This model also tended to collapse. Despite having more parameters than MUNIT and Img2Img SA, Pix2Pix was the next fastest model but took the longest number of epochs to converge. MUNIT took closer to two minutes for every thousand images, but could be trained in less epochs than Pix2Pix. With it's architecture of 2 generators and 2 discriminators, CycleGAN took almost three times as much time as Pix2Pix, but converged in a third of the epochs. Finally, both StarGAN2 and Img2Img SA were the slowest models and took over 500 seconds for every 1000 images. 

\begin{figure*}[!htb]
   \begin{minipage}{0.45\textwidth}
     \centering
     \includegraphics[width=1.0\linewidth]{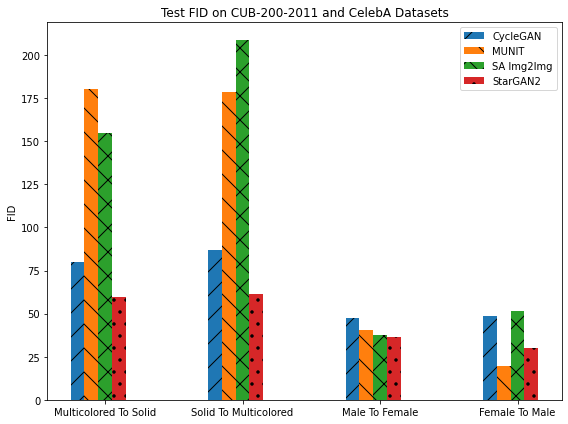}
     \caption{Test FID for each application on the CUB-200-2011 and CelebA datasets from Table~\ref{tab:restrans2}.}
     \label{fig:testfidtranstwo}
   \end{minipage}\hfill
   \begin{minipage}{0.45\textwidth}
     \centering
     \includegraphics[width=1.0\linewidth]{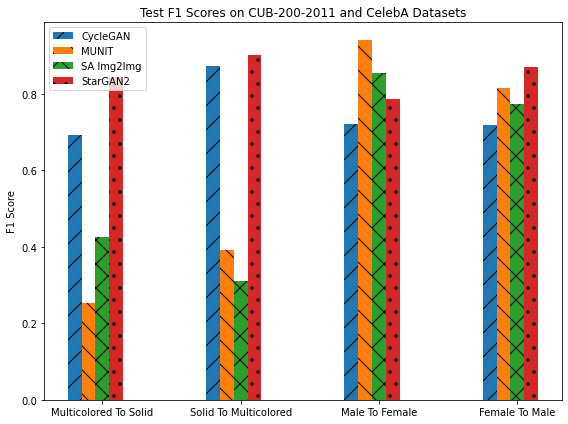}
     \caption{Test F1 Score for each application on the CUB-200-2011 and CelebA datasets from Table~\ref{tab:restrans2}.}
     \label{fig:testf1transtwo}
   \end{minipage}
\end{figure*}

For each dataset, an image translation was defined - $edge \leftrightarrow shoe$ for UT Zappos50k, $night \leftrightarrow day$ for Transient Attributes, $Solid\;Wings \leftrightarrow Multicolored\;Wings$ for CUB-200-2011, and $Male \leftrightarrow Female$ for CelebA. The results for these tasks are shown in Table~\ref{tab:restrans1}, Table~\ref{tab:restrans2}, and Figures~\ref{fig:edge2shoe} to~\ref{fig:female2male}. The two tables show the train and test FID, Precision, Recall, and F1-Score for each application the models were trained for and Figures~\ref{fig:testfidtransone} to~\ref{fig:testf1transtwo} visually depict the test FID and F1 scores. Figures~\ref{fig:edge2shoe} to~\ref{fig:female2male} show the translations of five randomly sampled images by each model. For Figures~\ref{fig:edge2shoe} to~\ref{fig:day2night}, the real paired output is also included. 

We were unable to reproduce the results presented in CoGAN using the PyTorch implementation provided in the official code base. For all methods, the model seemed to collapse, but this may be due to a misconfiguration of the hyperparameters. In Section~\ref{section:appendix}, we provide a small sample of our experiments using CoGAN. We also provide the results presented in the original paper which are a more accurate representation of the output that can be generated by CoGAN. 

\subsection{Analysis}

\begin{figure*}[!htb]
   \begin{minipage}{1.0\textwidth}
     \centering
     \includegraphics[width=1.0\linewidth]{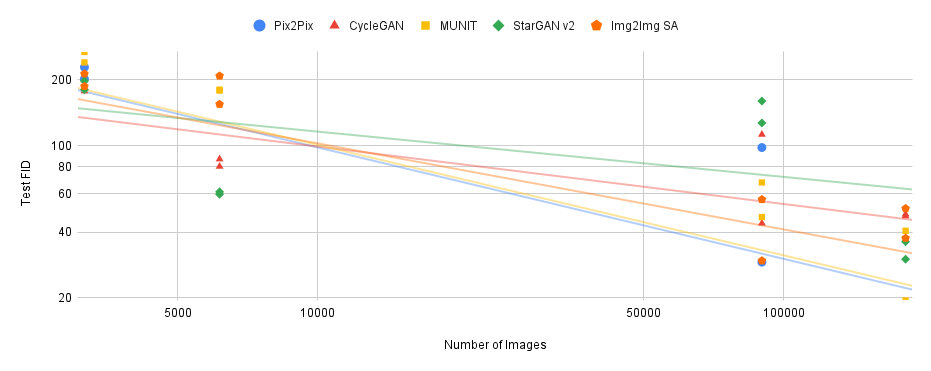}
     \caption{Test FID plotted against the number of images in the train dataset for each image translation task. The gridlines for both the Test FID and number of images have been marked on a log scale to allow for better visualization of the data. Given that there are only a maximum of 8 points for each model (2 for each dataset), the trendlines are solely meant to help better visualize the results.}
     \label{fig:testfidtransimgs}
   \end{minipage}
\end{figure*}

Overall, the performance of all four models was very dependent on the number of images present in the dataset. Figure~\ref{fig:testfidtransimgs} shows a very clear relationship between the number of images in the training dataset and the test FID for each model for the image translation tasks. Figure~\ref{fig:testfidtransimgs} also shows that despite training on the same dataset, all models achieve different FID scores depending on the direction of translation.

\begin{table*}[h]
  \scriptsize
  \begin{center}
    \caption{Results Of Image Translation on Transient Attributes and UT Zappos50k}
    \label{tab:restrans1}
    \sisetup{detect-weight,mode=text}
    \renewrobustcmd{\bfseries}{\fontseries{b}\selectfont}
    \renewrobustcmd{\boldmath}{}
    \newrobustcmd{\bs}{\bfseries}
    
    \npdecimalsign{.}
    \nprounddigits{3}
    \begin{tabular}{l S S S S S n{1}{3}}
      \textbf{Application} & \textbf{Model} & \textbf{Phase} & \textbf{FID} & \textbf{Precision} & \textbf{Recall} & \textbf{F1 Score}\\
      \hline
        \multirow[t]{12}{*}{Night To Day}
        & {\multirow[t]{2}{*}{Pix2Pix}} & {Train} & \bs 83.663000 & 0.726000 & \bs 0.787000 & \bs 0.755270\\
        & & {Test} & 202.146000 & 0.016000 & 0.067000 & 0.025831\\
        & {\multirow[t]{2}{*}{CycleGAN}} & {Train} & 93.603000 & 0.502000 & 0.395000 & 0.442118\\
        & & {Test} & 199.840000 & 0.079000 & 0.169000 & 0.107669\\
        & {\multirow[t]{2}{*}{MUNIT}} & {Train} & 177.982656 & 0.120000 & 0.091000 & 0.103507\\
        & & {Test} & 268.827108 & 0.000000 & 0.000000 & 0.000000\\
        & {\multirow[t]{2}{*}{SA Img2Img}} & {Train} & 139.748409 & 0.296000 & 0.365000 & 0.326899\\
        & & {Test} & 213.382232 & 0.170000 & 0.077000 & 0.105992\\
        & {\multirow[t]{2}{*}{StarGAN2}} & {Train} & 106.268288 & \bs 0.768000 & 0.571000 & 0.655008\\
        & & {Test} & \bs 197.913357 & \bs 0.211000 & \bs 0.215000 & \bs 0.212981\\
        \hline
        \multirow[t]{12}{*}{Day To Night}
        & {\multirow[t]{2}{*}{Pix2Pix}} & {Train} & 126.294000 & 0.373000 & 0.413000 & 0.391982\\
        & & {Test} & 228.808000 & 0.015000 & 0.010000 & 0.012000\\
        & {\multirow[t]{2}{*}{CycleGAN}} & {Train} & \bs 80.721000 & 0.486000 & 0.625000 & 0.546805\\
        & & {Test} & \bs 177.672000 & 0.008000 & 0.017000 & 0.010880\\
        & {\multirow[t]{2}{*}{MUNIT}} & {Train} & 181.237540 & 0.259000 & 0.302000 & 0.278852\\
        & & {Test} & 240.590093 & 0.042000 & 0.002000 & 0.003818\\
        & {\multirow[t]{2}{*}{SA Img2Img}} & {Train} & 130.886096 & 0.431000 & 0.362000 & 0.393498\\
        & & {Test} & 186.889825 & 0.020000 & 0.001000 & 0.001905\\
        & {\multirow[t]{2}{*}{StarGAN2}} & {Train} & 99.237776 & \bs 0.531000 & \bs 0.659000 & \bs 0.588116\\
        & & {Test} & 180.300756 & \bs 0.074000 & \bs 0.114000 & \bs 0.089745\\
        \hline
        \multirow[t]{12}{*}{Edge To Shoe}
        & {\multirow[t]{2}{*}{Pix2Pix}} & {Train} & \bs 25.514000 & \bs 0.904000 & 0.804000 & 0.851073\\
        & & {Test} & \bs 29.197000 & 0.882000 & 0.844000 & 0.862582\\
        & {\multirow[t]{2}{*}{CycleGAN}} & {Train} & 39.870000 & 0.812000 & 0.769000 & 0.789915\\
        & & {Test} & 44.036000 & 0.826000 & 0.747000 & 0.784516\\
        & {\multirow[t]{2}{*}{MUNIT}} & {Train} & 43.639703 & 0.783000 & 0.646000 & 0.707933\\
        & & {Test} & 46.987805 & 0.784000 & 0.629000 & 0.697999\\
        & {\multirow[t]{2}{*}{SA Img2Img}} & {Train} & 26.179320 & 0.867000 & \bs 0.843000 & \bs 0.854832\\
        & & {Test} & 29.654177 & \bs 0.883000 & \bs 0.873000 & \bs 0.877972\\
        & {\multirow[t]{2}{*}{StarGAN2}} & {Train} & 157.517958 & 0.161000 & 0.608000 & 0.254585\\
        & & {Test} & 160.044674 & 0.167000 & 0.614000 & 0.262581\\
        \hline
        \multirow[t]{12}{*}{Shoe To Edge}
        & {\multirow[t]{2}{*}{Pix2Pix}} & {Train} & 93.513000 & 0.273000 & 0.464000 & 0.343750\\
        & & {Test} & 97.939000 & 0.262000 & 0.380000 & 0.310156\\
        & {\multirow[t]{2}{*}{CycleGAN}} & {Train} & 108.599000 & 0.592000 & 0.341000 & 0.432737\\
        & & {Test} & 112.408000 & 0.563000 & 0.348000 & 0.430130\\
        & {\multirow[t]{2}{*}{MUNIT}} & {Train} & 63.927897 & 0.609000 & 0.649000 & 0.628364\\
        & & {Test} & 67.709222 & 0.557000 & 0.591000 & 0.573497\\
        & {\multirow[t]{2}{*}{SA Img2Img}} & {Train} & \bs 52.789539 & \bs 0.623000 & \bs 0.679000 & \bs 0.649796\\
        & & {Test} & \bs 56.601596 & \bs 0.621000 & \bs 0.653000 & \bs 0.636598\\
        & {\multirow[t]{2}{*}{StarGAN2}} & {Train} & 122.080475 & 0.277000 & 0.335000 & 0.303252\\
        & & {Test} & 126.938574 & 0.288000 & 0.334000 & 0.309299\\
    \end{tabular}
  \end{center}
\end{table*}

\begin{table*}[h]
  \scriptsize
  \begin{center}
    \caption{Results Of Image Translation on CUB-200-2011 and CelebA}
    
    \sisetup{detect-weight,mode=text}
    \renewrobustcmd{\bfseries}{\fontseries{b}\selectfont}
    \renewrobustcmd{\boldmath}{}
    \newrobustcmd{\bs}{\bfseries}
    
    \label{tab:restrans2}
    
    \npdecimalsign{.}
    \nprounddigits{3}
    \begin{tabular}{l S S S S S n{1}{3}}
      \textbf{Application} & \textbf{Model} & \textbf{Phase} & \textbf{FID} & \textbf{Precision} & \textbf{Recall} & \textbf{F1 Score}\\
      \hline
        \multirow[t]{10}{*}{Multicolored}
        & {\multirow[t]{2}{*}{CycleGAN}} & {Train} & 56.910000 & 0.752000 & 0.723000 & 0.737215\\
        To Solid & & {Test} & 80.221000 & 0.766000 & 0.630000 & 0.691375\\
        & {\multirow[t]{2}{*}{MUNIT}} & {Train} & 133.862267 & 0.550000 & 0.295000 & 0.384024\\
        & & {Test} & 180.489803 & 0.524000 & 0.168000 & 0.254428\\
        & {\multirow[t]{2}{*}{SA Img2Img}} & {Train} & 104.121065 & 0.698000 & 0.415000 & 0.520521\\
        & & {Test} & 154.647318 & 0.631000 & 0.321000 & 0.425527\\
        & {\multirow[t]{2}{*}{StarGAN2}} & {Train} & \bs 53.998975 & \bs 0.831000 & \bs 0.915000 & \bs 0.870979\\
        & & {Test} & \bs 59.805613 & \bs 0.821000 & \bs 0.868000 & \bs 0.843846\\
        \hline
        \multirow[t]{10}{*}{Solid To}
        & {\multirow[t]{2}{*}{CycleGAN}} & {Train} & \bs 44.394000 & 0.669000 & 0.780000 & 0.720248\\
        Multicolored & & {Test} & 86.704000 & 0.947000 & 0.809000 & 0.872577\\
        & {\multirow[t]{2}{*}{MUNIT}} & {Train} & 140.702892 & 0.469000 & 0.281000 & 0.351437\\
        & & {Test} & 178.719411 & 0.477000 & 0.333000 & 0.392200\\
        & {\multirow[t]{2}{*}{SA Img2Img}} & {Train} & 154.930285 & 0.487000 & 0.240000 & 0.321541\\
        & & {Test} & 208.582341 & 0.417000 & 0.246000 & 0.309448\\
        & {\multirow[t]{2}{*}{StarGAN2}} & {Train} & 52.887697 & \bs 0.916000 & \bs 0.840000 & \bs 0.876355\\
        & & {Test} & \bs 61.337908 & \bs 0.907000 & \bs 0.896000 & \bs 0.901466\\
        \hline
        \multirow[t]{10}{*}{Male To Female}
        & {\multirow[t]{2}{*}{CycleGAN}} & {Train} & 45.117000 & 0.736000 & 0.721000 & 0.728423\\
        & & {Test} & 47.529000 & 0.745000 & 0.700000 & 0.721799\\
        & {\multirow[t]{2}{*}{MUNIT}} & {Train} & 38.640336 & \bs 0.936000 & \bs 0.933000 & \bs 0.934498\\
        & & {Test} & 40.630407 & \bs 0.947000 & \bs 0.933000 & \bs 0.939948\\
        & {\multirow[t]{2}{*}{SA Img2Img}} & {Train} & 35.192499 & 0.821000 & 0.856000 & 0.838135\\
        & & {Test} & 37.615791 & 0.832000 & 0.877000 & 0.853908\\
        & {\multirow[t]{2}{*}{StarGAN2}} & {Train} & \bs 34.328942 & 0.728000 & 0.835000 & 0.777837\\
        & & {Test} & \bs 36.249056 & 0.744000 & 0.836000 & 0.787322\\
        \hline
        \multirow[t]{10}{*}{Female To Male}
        & {\multirow[t]{2}{*}{CycleGAN}} & {Train} & 45.786000 & 0.662000 & 0.734000 & 0.696143\\
        & & {Test} & 48.420000 & 0.709000 & 0.727000 & 0.717887\\
        & {\multirow[t]{2}{*}{MUNIT}} & {Train} & \bs 17.952819 & 0.760000 & 0.883000 & 0.816896\\
        & & {Test} & \bs 19.486343 & 0.757000 & 0.882000 & 0.814733\\
        & {\multirow[t]{2}{*}{SA Img2Img}} & {Train} & 49.186183 & 0.718000 & 0.811000 & 0.761672\\
        & & {Test} & 51.459449 & 0.732000 & 0.819000 & 0.773060\\
        & {\multirow[t]{2}{*}{StarGAN2}} & {Train} & 27.614688 & \bs 0.810000 & \bs 0.945000 & \bs 0.872308\\
        & & {Test} & 30.120232 & \bs 0.816000 & \bs 0.934000 & \bs 0.871022\\
    \end{tabular}
  \end{center}
\end{table*}

\begin{figure*}[!htb]
   \begin{minipage}{0.45\textwidth}
     \centering
     \includegraphics[width=1.0\linewidth]{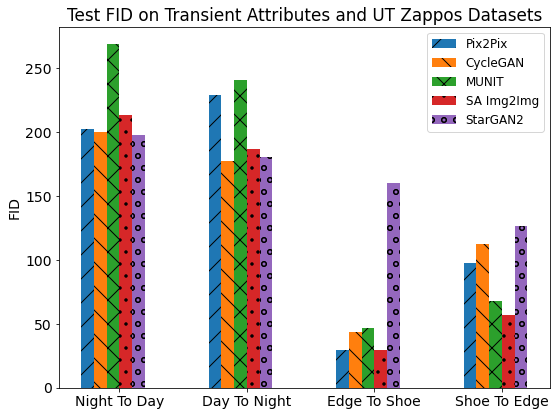}
     \caption{Test FID for each application on the UT Zappos50k and Transient Attributes datasets from Table~\ref{tab:restrans1}.}
     \label{fig:testfidtransone}
   \end{minipage}\hfill
   \begin{minipage}{0.45\textwidth}
     \centering
     \includegraphics[width=1.0\linewidth]{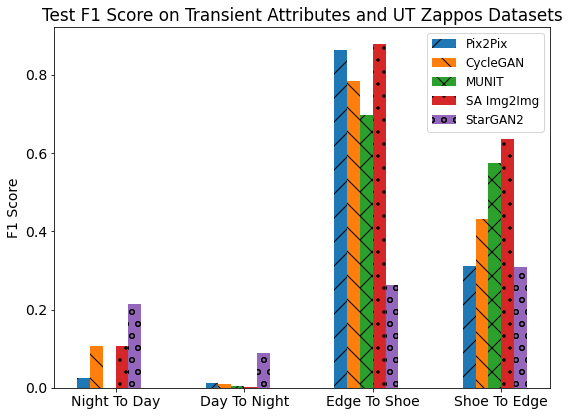}
     \caption{Test F1 Scores for each application on the UT Zappos50k and Transient Attributes datasets from Table~\ref{tab:restrans1}.}
     \label{fig:testf1transone}
   \end{minipage}\hfill
\end{figure*}

For the image translation tasks with paired data (Transient Attributes and UT Zappos datasets on Table~\ref{tab:restrans1}), we may have expected Pix2Pix to outperform all other methods given that it can take advantage of pair-wise supervision. Instead, in Figure~\ref{fig:testfidtransone}, we see that all other methods outperform Pix2Pix on at least one task and that Pix2Pix only outperformed most methods on the $Edge \rightarrow Shoe$ task. We can see the same trends in the F1 score for each model shown in Figure~\ref{fig:testf1transone} with Pix2Pix scoring relatively low F1 scores for all tasks except $Edge \rightarrow Shoe$. This seems to indicate that pair-level supervision is not necessary to achieve state-of-the-art results, and in some cases may actually hinder the model from generalizing to the test set. However, this does not mean that pair-level supervision does not improve model performance. We would need an ablation study on a model that combines the advancements in each model to be able to determine the effect that pair-level supervision has on model performance, but that is beyond the scope of this paper and remains a task for future researchers.

In all tasks with paired image data, Img2Img SA consistently achieves comparable or superior FID scores. Expanding to all tasks and including the results shown in Table~\ref{tab:restrans2}, we can see that both Img2Img SA and StarGAN2 present some of the best numerical results in all datasets with a few key exceptions. StarGAN2 underperforms on the UT Zappos50k dataset and Img2Img SA underperforms on the CUB-200-2011 dataset. However, in both cases, we see that the other model presents the best FID or F1 scores. CycleGAN and MUNIT don't seem to perform as well across all tasks, but present competitive performance. MUNIT has the lowest FID Score on the $Female \rightarrow Male$ task and the highest F1 Score on $Male \rightarrow Female$ task which may indicate that given enough training images, MUNIT may have the potential to outperform both StarGAN2 and Img2Img SA. 

\begin{figure*}[!htb]
   \begin{minipage}{0.45\textwidth}
     \centering
     \includegraphics[width=1.0\linewidth]{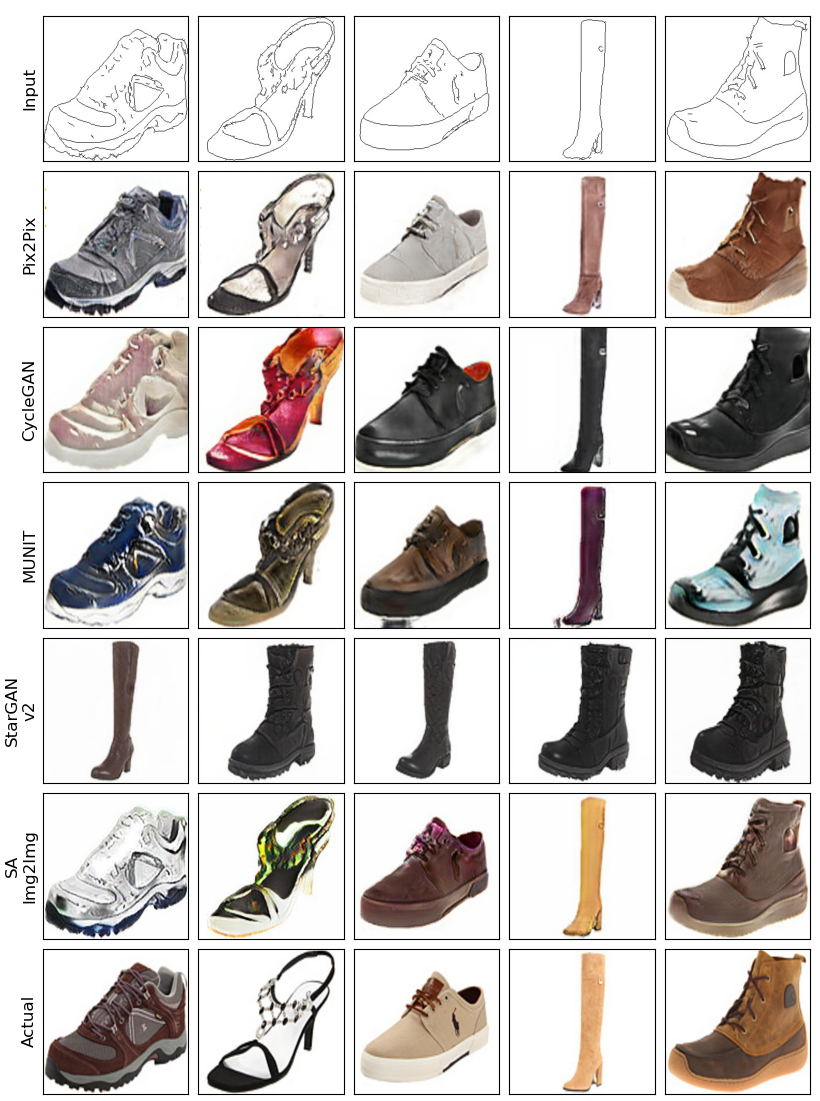}
     \caption{Results of Edge To Shoe Image Translation on UT Zappos50k Test Dataset}\label{fig:edge2shoe}
   \end{minipage}\hfill
   \begin{minipage}{0.45\textwidth}
     \centering
     \includegraphics[width=1.0\linewidth]{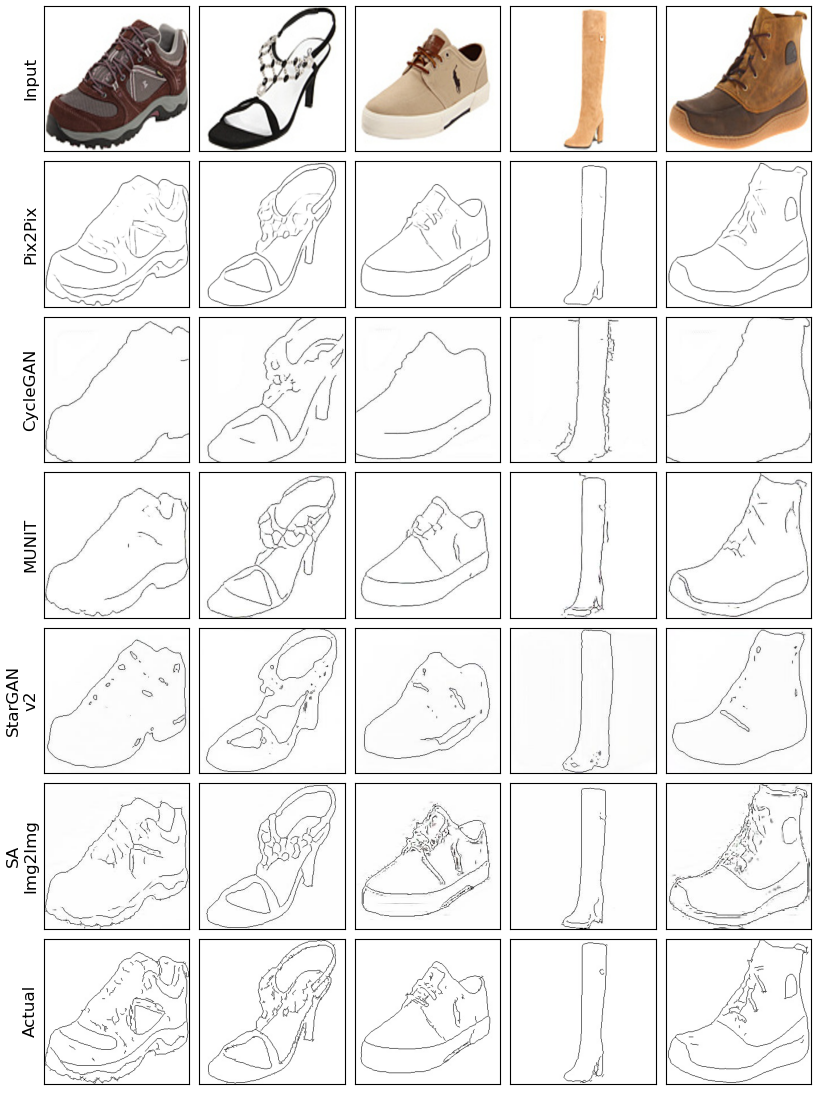}
     \caption{Results of Shoe To Edge Image Translation on UT Zappos50k Test Dataset}\label{fig:shoe2edge}
   \end{minipage}
\end{figure*}

\begin{figure*}[!htb]
   \begin{minipage}{0.48\textwidth}
     \centering
     \includegraphics[width=1.0\linewidth]{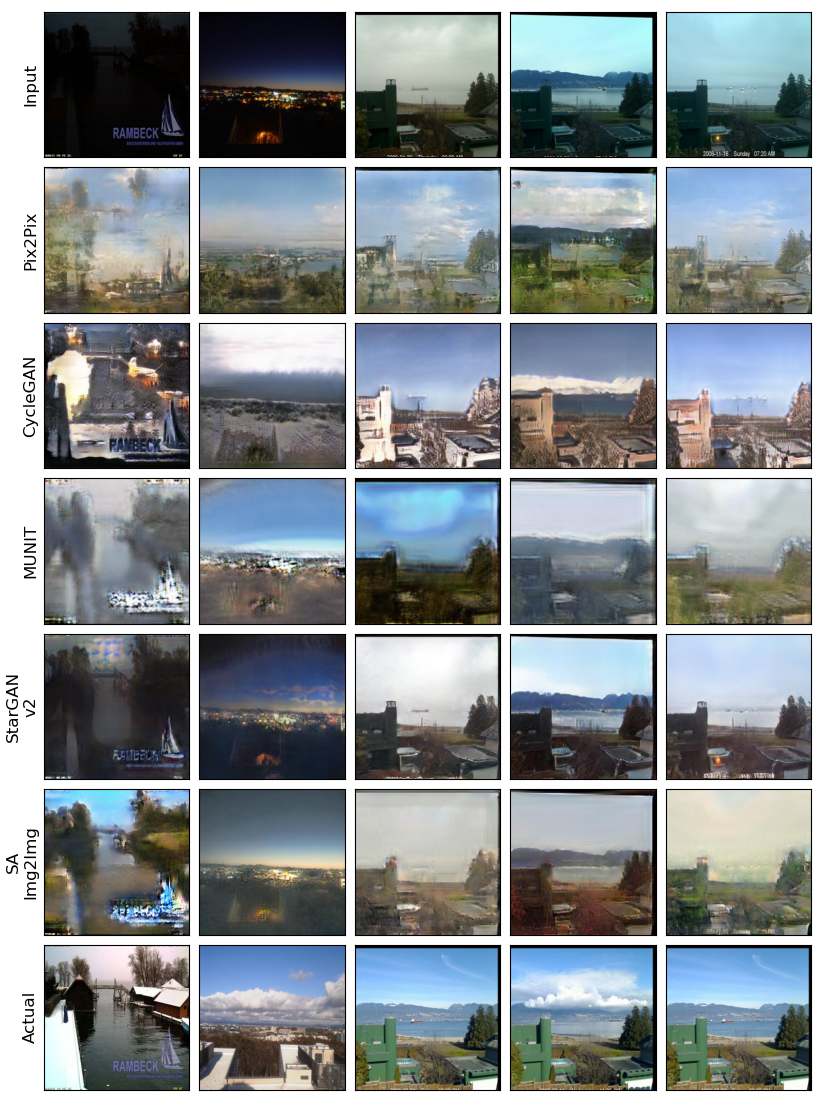}
     \caption{Results of Night To Day Image Translation on Transient Attributes Test Dataset}\label{fig:night2day}
   \end{minipage}\hfill
   \begin{minipage}{0.48\textwidth}
     \centering
     \includegraphics[width=1.0\linewidth]{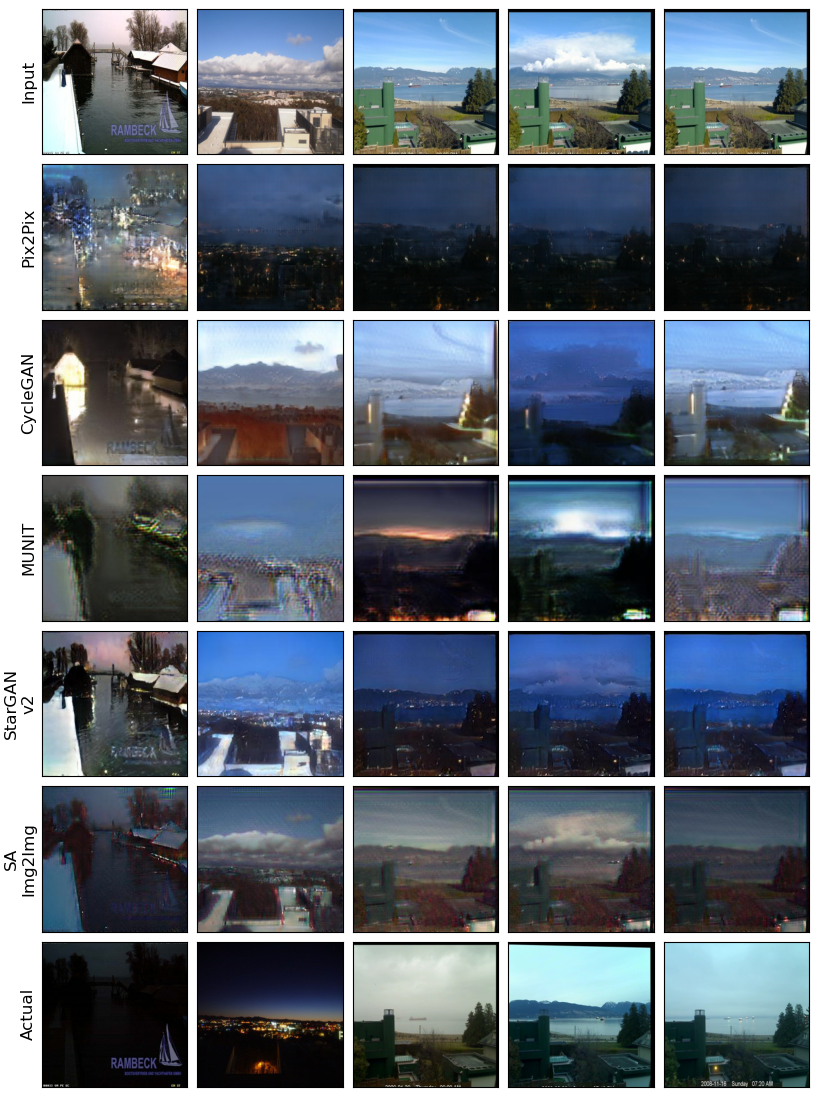}
     \caption{Results of Day To Night Image Translation on Transient Attributes Test Dataset}\label{fig:day2night}
   \end{minipage}
\end{figure*}

\begin{figure*}[!htbp]
   \begin{minipage}{0.48\textwidth}
     \centering
     \includegraphics[width=1.0\linewidth]{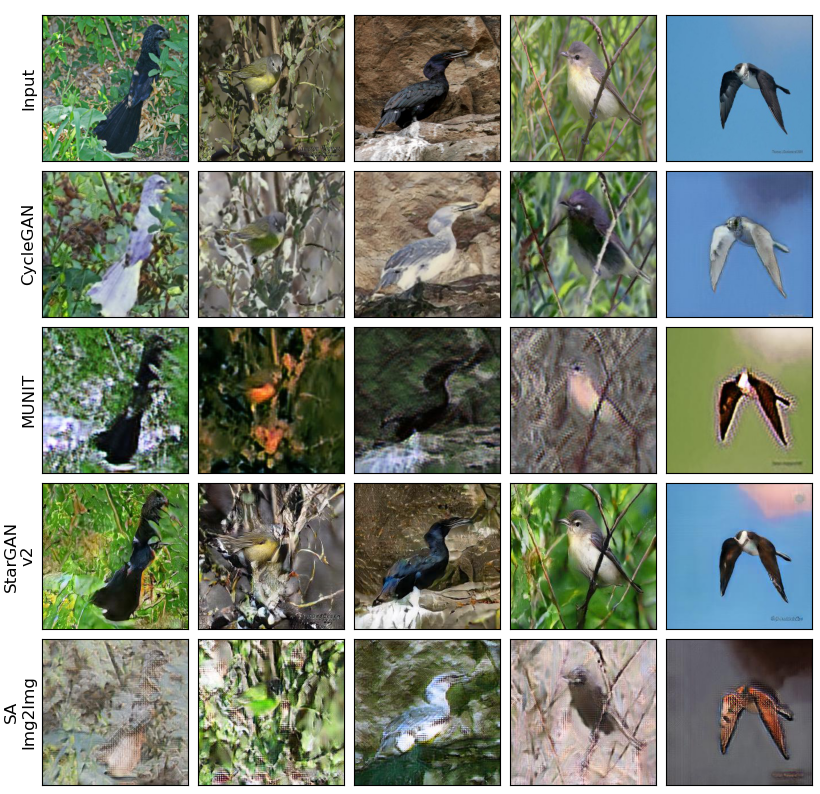}
     \caption{Results of Solid Wing To Multicolored Wing Translation on CUB-200-2011 Test Dataset}\label{fig:solid2mc}
   \end{minipage}\hfill
   \begin{minipage}{0.48\textwidth}
     \centering
     \includegraphics[width=1.0\linewidth]{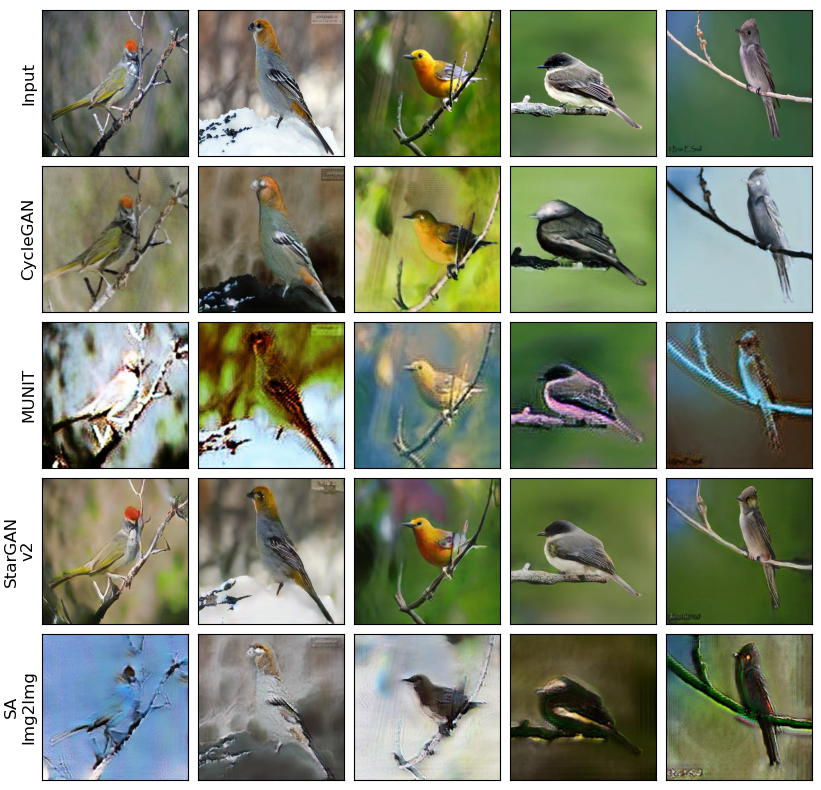}
     \caption{Results of Multicolored Wing To Solid Wing Image Translation on CUB-200-2011 Test Dataset}\label{fig:mc2solid}
   \end{minipage}
\end{figure*}

\begin{figure*}[!htbp]
   \begin{minipage}{0.45\textwidth}
     \centering
     \includegraphics[width=1.0\linewidth]{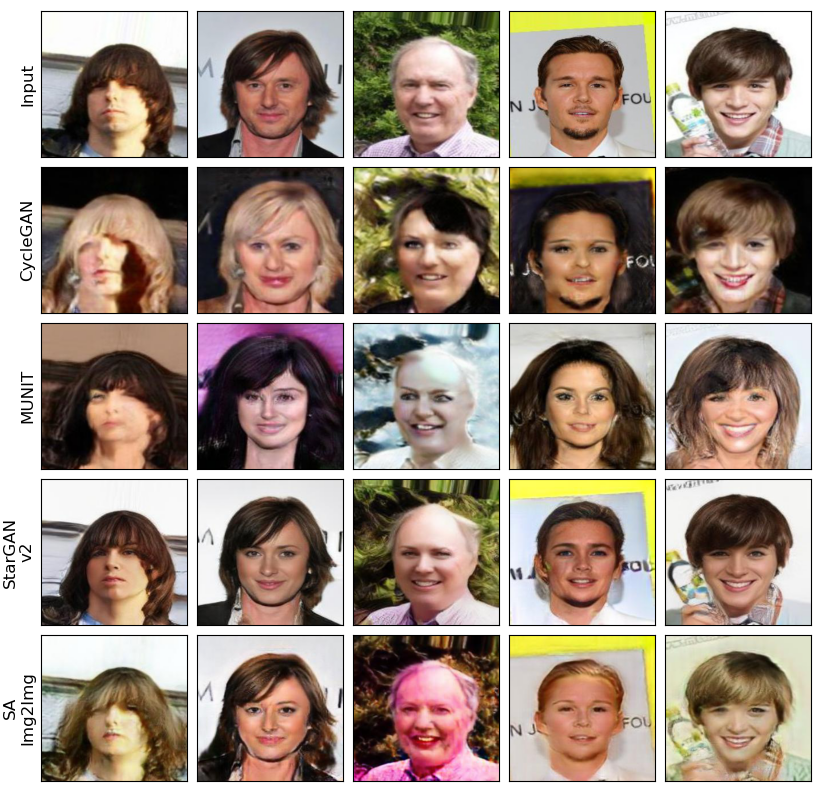}
     \caption{Results of Male To Female Image Translation on CelebA Test Dataset}\label{fig:male2female}
   \end{minipage}\hfill
   \begin{minipage}{0.45\textwidth}
     \centering
     \includegraphics[width=1.0\linewidth]{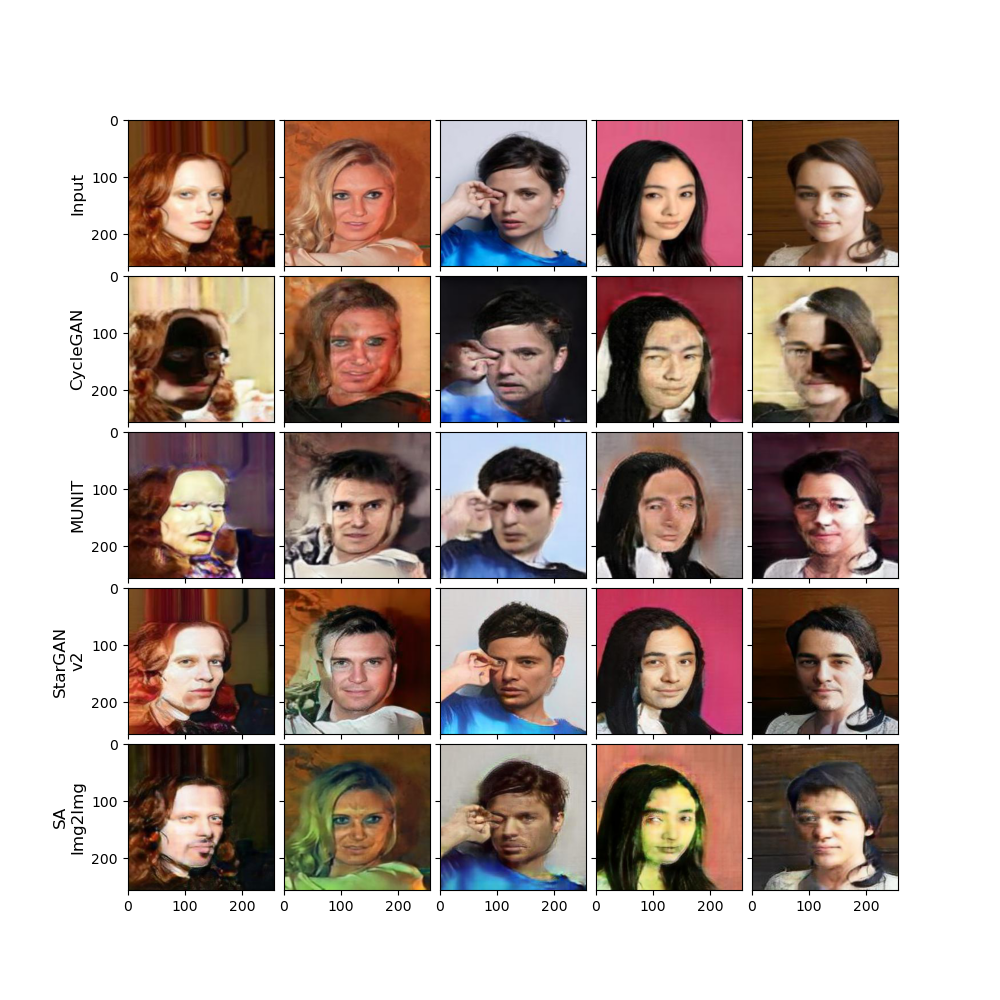}
     \caption{Results of Female To Male Image Translation on CelebA Test Dataset}\label{fig:female2male}
   \end{minipage}
\end{figure*}

The visual results in figures~\ref{fig:edge2shoe} to~\ref{fig:female2male} give more insight into how the models performed. The FID and F1 Score indicated that StarGAN2 performed poorly on the UT Zappos50k dataset; in Figure~\ref{fig:edge2shoe}, it's clear that StarGAN2 is creating similar shoes for any given input edge which may indicate mode collapse. The shoes also don't have the same level of variety in style as the ones generated by all other GANs (most shoes tend to be black). For the generated edges in Figure~\ref{fig:shoe2edge}, both CycleGAN and StarGAN2 also lack the same level of detail that appears in other generated and real edges.

Similarly, the FID and F1 Score indicated that both MUNIT and SA Img2Img performed poorly on the CUB200 dataset. Both methods were outperformed by StarGAN2 and CycleGAN. StarGAN2 was able to produce high quality outputs by generating a brighter version of the original image for both $solid \rightarrow multicolored$ and $multicolored \rightarrow solid$ tasks. Despite their FID and F1 Scores being worse than StarGAN2, MUNIT and SA Img2Img may arguably be superior methods on the CUB200 dataset as they learned more interesting translations rather than the identity of an input image. CycleGAN was also outperformed by StarGAN2, but it may have generated the best sample images by performing color transformations on the birds and only slightly modifying the background. 

The visual results in Figure~\ref{fig:night2day} and Figure~\ref{fig:day2night} also present interesting results in the deviation between numerical metrics and visual output. Due to the small size of the dataset, 3 of the randomly sampled images were from the same scene. This is especially true for the $day \rightarrow night$ translation, where three of the input images are almost identical. Figure~\ref{fig:day2night} shows that although StarGAN2 and SA Img2Img present higher quality images, MUNIT presents a more diverse set of output images. This conclusion isn't apparent from the F1 Score, which presents a higher F1 Score for StarGAN2 (despite an almost deterministic output). The output quality of the images may be one reason the diversity in the generated images of MUNIT do not translate to the F1 Score.
 
It's important to note that these conclusions are being derived from a random sample of five. A random sample of five is not representative of the entire dataset, but these results show that although FID, precision, and recall alone may be good benchmarks for the performance of these GANs, they should be complimented by other measures from Section~\ref{section:metrics} to give a more complete picture of the performance of models. 

A key challenge of designing these experiments was choosing the right attributes to translate on the data. On the paired datasets, all models were able to learn the distinction between edge and shoe for the UT Zappos50k dataset and day and night for the Transient Attributes dataset relatively well. Most models presented results comparable or better than Pix2Pix. The results on the CUB-200-2011 and CelebA datasets were less promising. As noted previously, Figures~\ref{fig:solid2mc} and~\ref{fig:mc2solid} show that both CycleGAN and SA Img2Img learned to invert the colors of the birds and the backgrounds, MUNIT added color transformations, and StarGAN2 presented a brightened identity transformation. Figures~\ref{fig:male2female} and ~\ref{fig:female2male} show that CycleGAN, SA Img2Img, and MUNIT chose to change both facial attributes and the background when translating between domains and SA Img2Img also changed skin tone for some of the images.

Since image-to-image GANs are unsupervised models, it is always possible that these models learn something completely different about two image distributions. One advantage that StarGAN2 has over other methods is its domain-specific discriminator. The domain-specific discriminator allows StarGAN to ensure that images are not only real or fake, but also belong to the domain they have been translated to. This may have been one of the reasons that the sample of StarGAN's translations on the CelebA dataset seem more realistic. 

For the CUB-200-2011 dataset, however, it is difficult for even human observers to note that the difference between the two sets of birds are the patterns of the wings. Even with self-attention mechanisms in Img2Img SA, it's clear that just the image distributions of birds with solid wings and birds with multicolored images are insufficient to model that task. To perform image-to-image translation with that level of detail, additional input would need to be provided to the model.

\section{Conclusion}

In this paper, we compare each model's architecture, loss functions, and the new techniques that they introduced for Image-to-Image translation. More specifically, we

\begin{itemize} 
 \item{present a survey of 8 Image-to-Image GANs,}
 \item{survey the 9 metrics and 18 datasets that these GANs were trained on,}
 \item{and analyze 6 of these Image-to-Image GANs on a common set of datasets and metrics}
\end{itemize}
 
Our results showed that no model is perfect for all applications and datasets. Pix2Pix performed well with paired data on large datasets, but didn't generalize as well with smaller datasets. CycleGAN performed consistently on all applications but was always outperformed by either StarGAN2 or SA Img2Img. MUNIT produced diverse outputs given similar input images, but tended to generate lower quality images and was not as consistent as CycleGAN. SA Img2Img and StarGAN2 alternated in outperforming other methods and produced high-quality images when they did, but also tended to underperform all other methods on certain tasks. At their core, however, these models are all performing Image-to-Image translation. Choosing different models for different applications and datasets seems to go against the original concept of automatic image-to-image translation proposed with Pix2Pix by Isola et al. 

In addition, the summaries of each metric described in Section~\ref{section:metrics} can help future researchers decide which metrics are the best ones to choose for their specific model. Table~\ref{tab:datadesc} on the descriptions of datasets can help future researchers decide which datasets they should train their models on. Finally, our analysis and discussion will allow researchers to compare the performance of these models and the techniques prior to implementing similar techniques for their own models. 

As researchers create new Image-to-Image GANs, some future directions to pursue could be efficiently training Image-to-Image GANs with less computational resources, improving performance on small datasets, choosing between domain level or instance-level supervision, choosing the right attributes, balancing image quality and diversity, choosing the right metrics to gain a comprehensive view of the results, and selecting the right datasets to prove that a model could be extrapolated to other datasets and domains.

\bigskip

\begin{appendices}

\section{CoGAN and DA-GAN Results}
\label{section:appendix}

In this section, we include the results of CoGAN that we had found by running our experiments and the original results from the CoGAN paper. Since researchers had not published their code, we also include some of the original visual results from Deep Attention GAN in this section.

\subsection{CoGAN}

For all datasets, we were unable to train CoGAN without it collapsing using the official PyTorch implementation provided. These results deviated significantly from the results presented in the original paper, and there may be some issues in the setup of the official repository, incorrect configuration of certain hyperparameters, or missing steps of fine tuning required to generalize to new datasets.

\begin{table}[h!]
  \begin{center}
    \caption{CoGAN Test Results}
    \label{tab:coganmetrics}
    \begin{tabular}{l S S S r}
      \textbf{Task} & \textbf{FID} & \textbf{Precision} & \textbf{Recall} & \textbf{F1}\\
      \hline
      {Night To Day}          & {411.676} & {0} & {0} & {0}\\
      {Day To Night}          & {369.807} & {0} & {0} & {0}\\
      {Edge To Shoe}          & {345.071} & {0} & {0} & {0}\\
      {Shoe To Edge}          & {440.493} & {0} & {0} & {0}\\
      {Solid To Multicolored} & {314.376} & {0} & {0} & {0}\\
      {Multicolored To Solid} & {313.118} & {0} & {0} & {0}\\
    \end{tabular}
  \end{center}
\end{table}

\begin{figure}[!htbp]
   \centering
   \begin{minipage}{0.4\textwidth}
     \centering
     \includegraphics[width=1.0\linewidth]{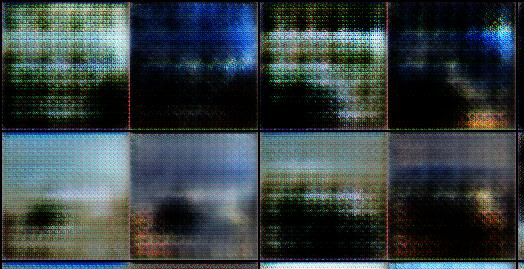}
     \caption{Sample generated results on the night to day image translation task}\label{fig:cogan-output}
   \end{minipage}
\end{figure}

Table~\ref{tab:coganmetrics} demonstrates that for all datasets, the FID score was over 300 and the precision and recall for each dataset was 0. This was a clear indication that the model was unable to generate high-quality output. This was also corroborated with the visual results of CoGAN (as shown by a sample 
in Figure~\ref{fig:cogan-output}).

\begin{figure}[!htbp]
   \centering
   \begin{minipage}{0.5\textwidth}
     \centering
     \includegraphics[width=1.0\linewidth]{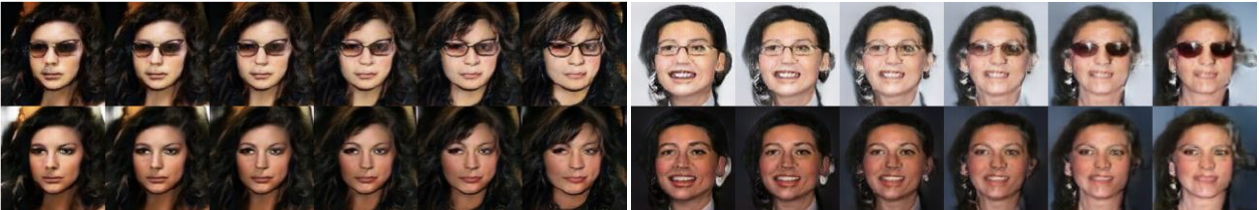}
     \caption{Original results of paired image generation presented in Figure 4 in \cite{cogan}}\label{fig:cogan-pairs}
   \end{minipage}
\end{figure}

The visual results from \cite{cogan} in Figure~\ref{fig:cogan-pairs} show that this model is able to generate high-quality output, especially when generating paired images for a standard image generation task.

\begin{figure}[!htbp]
   \centering
   \begin{minipage}{0.4\textwidth}
     \centering
     \includegraphics[width=1.0\linewidth]{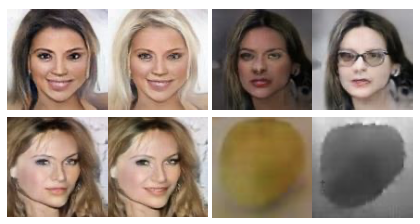}
     \caption{Original results of cross-domain image translation presented in Figure 6 in \cite{cogan}}\label{fig:cogan-cross-domain}
   \end{minipage}
\end{figure}

The original results in \cite{cogan} for cross-domain image translation, presented high-quality image translations between two domains in most cases. However, in some cases, the model was unable to find a good noise vector to generate a close pair to a given image (as shown in the bottom right example in Figure~\ref{fig:cogan-cross-domain}).

\subsection{Deep Attention GAN}

\begin{figure}[!htbp]
   \centering
   \begin{minipage}{0.45\textwidth}
     \centering
     \includegraphics[width=0.5\linewidth]{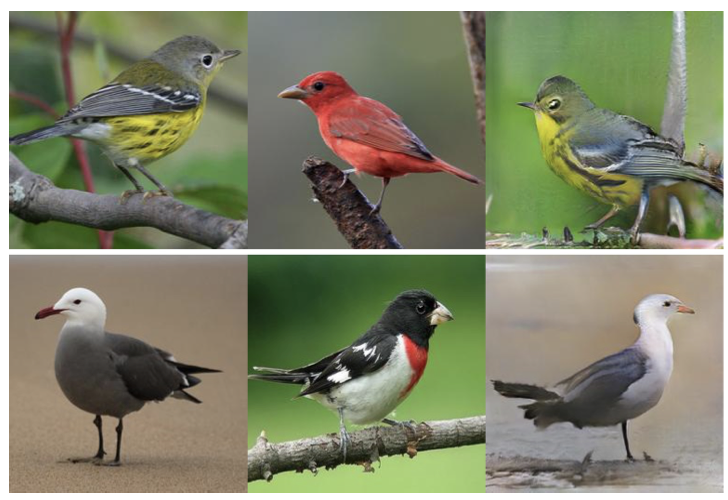}
     \caption{Original results of pose-morphing with image translation in Figure 7 in \cite{dagan}}\label{fig:dagan-pose-morph}
   \end{minipage}\hfill
   \begin{minipage}{0.45\textwidth}
     \centering
     \includegraphics[width=0.5\linewidth]{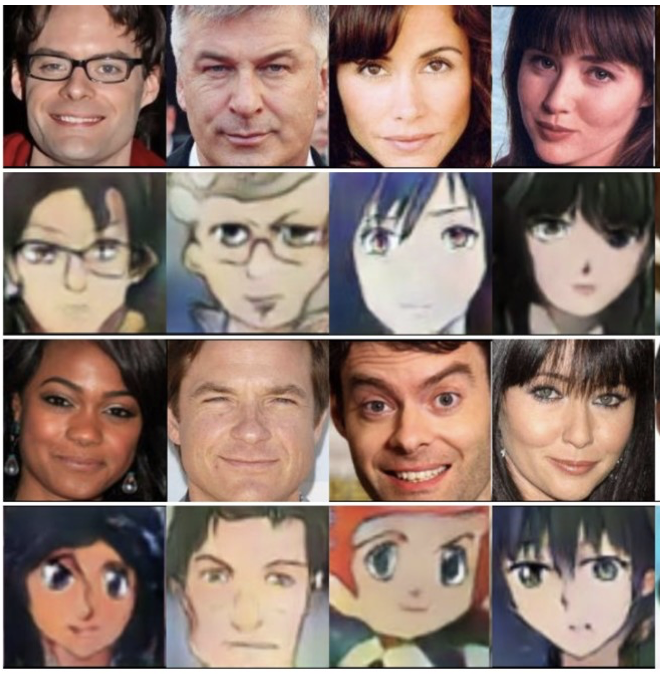}
     \caption{Original results of human face to animated face image translation in Figure 8b in \cite{dagan}}\label{fig:dagan-face-trans}
   \end{minipage}
\end{figure}

\section{Data Availability}

All datasets used in this paper are public datasets and have been cited and described in Table~\ref{tab:datadesc}. If the exact train-test splits are required, data is available on request from the authors.

\end{appendices}


\bibliography{bibliography}


\begin{thebibliography}{53}
\ifx \bisbn   \undefined \def \bisbn  #1{ISBN #1}\fi
\ifx \binits  \undefined \def \binits#1{#1}\fi
\ifx \bauthor  \undefined \def \bauthor#1{#1}\fi
\ifx \batitle  \undefined \def \batitle#1{#1}\fi
\ifx \bjtitle  \undefined \def \bjtitle#1{#1}\fi
\ifx \bvolume  \undefined \def \bvolume#1{\textbf{#1}}\fi
\ifx \byear  \undefined \def \byear#1{#1}\fi
\ifx \bissue  \undefined \def \bissue#1{#1}\fi
\ifx \bfpage  \undefined \def \bfpage#1{#1}\fi
\ifx \blpage  \undefined \def \blpage #1{#1}\fi
\ifx \burl  \undefined \def \burl#1{\textsf{#1}}\fi
\ifx \doiurl  \undefined \def \doiurl#1{\url{https://doi.org/#1}}\fi
\ifx \betal  \undefined \def \betal{\textit{et al.}}\fi
\ifx \binstitute  \undefined \def \binstitute#1{#1}\fi
\ifx \binstitutionaled  \undefined \def \binstitutionaled#1{#1}\fi
\ifx \bctitle  \undefined \def \bctitle#1{#1}\fi
\ifx \beditor  \undefined \def \beditor#1{#1}\fi
\ifx \bpublisher  \undefined \def \bpublisher#1{#1}\fi
\ifx \bbtitle  \undefined \def \bbtitle#1{#1}\fi
\ifx \bedition  \undefined \def \bedition#1{#1}\fi
\ifx \bseriesno  \undefined \def \bseriesno#1{#1}\fi
\ifx \blocation  \undefined \def \blocation#1{#1}\fi
\ifx \bsertitle  \undefined \def \bsertitle#1{#1}\fi
\ifx \bsnm \undefined \def \bsnm#1{#1}\fi
\ifx \bsuffix \undefined \def \bsuffix#1{#1}\fi
\ifx \bparticle \undefined \def \bparticle#1{#1}\fi
\ifx \barticle \undefined \def \barticle#1{#1}\fi
\bibcommenthead
\ifx \bconfdate \undefined \def \bconfdate #1{#1}\fi
\ifx \botherref \undefined \def \botherref #1{#1}\fi
\ifx \url \undefined \def \url#1{\textsf{#1}}\fi
\ifx \bchapter \undefined \def \bchapter#1{#1}\fi
\ifx \bbook \undefined \def \bbook#1{#1}\fi
\ifx \bcomment \undefined \def \bcomment#1{#1}\fi
\ifx \oauthor \undefined \def \oauthor#1{#1}\fi
\ifx \citeauthoryear \undefined \def \citeauthoryear#1{#1}\fi
\ifx \endbibitem  \undefined \def \endbibitem {}\fi
\ifx \bconflocation  \undefined \def \bconflocation#1{#1}\fi
\ifx \arxivurl  \undefined \def \arxivurl#1{\textsf{#1}}\fi
\csname PreBibitemsHook\endcsname

\bibitem{gan}
\begin{bchapter}
\bauthor{\bsnm{Goodfellow}, \binits{I.J.}},
\bauthor{\bsnm{Pouget-Abadie}, \binits{J.}},
\bauthor{\bsnm{Mirza}, \binits{M.}},
\bauthor{\bsnm{Xu}, \binits{B.}},
\bauthor{\bsnm{Warde-Farley}, \binits{D.}},
\bauthor{\bsnm{Ozair}, \binits{S.}},
\bauthor{\bsnm{Courville}, \binits{A.}},
\bauthor{\bsnm{Bengio}, \binits{Y.}}:
\bctitle{Generative adversarial nets}.
In: \bbtitle{Proceedings of the 27th International Conference on Neural
  Information Processing Systems - Volume 2}.
\bsertitle{NIPS’14},
pp. \bfpage{2672}--\blpage{2680}.
\bpublisher{MIT Press},
\blocation{Cambridge, MA, USA}
(\byear{2014})
\end{bchapter}
\endbibitem

\bibitem{styletrans}
\begin{bchapter}
\bauthor{\bsnm{Li}, \binits{C.}},
\bauthor{\bsnm{Wand}, \binits{M.}}:
\bctitle{Precomputed real-time texture synthesis with markovian generative
  adversarial networks}.
In: \bbtitle{Computer Vision - {ECCV} 2016 - 14th European Conference,
  Amsterdam, The Netherlands, October 11-14, 2016, Proceedings, Part {III}}.
\bsertitle{Lecture Notes in Computer Science},
vol. \bseriesno{9907},
pp. \bfpage{702}--\blpage{716}
(\byear{2016})
\end{bchapter}
\endbibitem

\bibitem{stylegan}
\begin{bchapter}
\bauthor{\bsnm{{Karras}}, \binits{T.}},
\bauthor{\bsnm{{Laine}}, \binits{S.}},
\bauthor{\bsnm{{Aila}}, \binits{T.}}:
\bctitle{A style-based generator architecture for generative adversarial
  networks}.
In: \bbtitle{2019 IEEE/CVF Conference on Computer Vision and Pattern
  Recognition (CVPR)},
pp. \bfpage{4396}--\blpage{4405}
(\byear{2019})
\end{bchapter}
\endbibitem

\bibitem{stylegan2}
\begin{bchapter}
\bauthor{\bsnm{Karras}, \binits{T.}},
\bauthor{\bsnm{Laine}, \binits{S.}},
\bauthor{\bsnm{Aittala}, \binits{M.}},
\bauthor{\bsnm{Hellsten}, \binits{J.}},
\bauthor{\bsnm{Lehtinen}, \binits{J.}},
\bauthor{\bsnm{Aila}, \binits{T.}}:
\bctitle{Analyzing and improving the image quality of stylegan}.
In: \bbtitle{The IEEE/CVF Conference on Computer Vision and Pattern Recognition
  (CVPR)},
pp. \bfpage{8110}--\blpage{8119}
(\byear{2020})
\end{bchapter}
\endbibitem

\bibitem{pix2pix}
\begin{bchapter}
\bauthor{\bsnm{{Isola}}, \binits{P.}},
\bauthor{\bsnm{{Zhu}}, \binits{J.}},
\bauthor{\bsnm{{Zhou}}, \binits{T.}},
\bauthor{\bsnm{{Efros}}, \binits{A.A.}}:
\bctitle{Image-to-image translation with conditional adversarial networks}.
In: \bbtitle{2017 IEEE Conference on Computer Vision and Pattern Recognition
  (CVPR)},
pp. \bfpage{5967}--\blpage{5976}
(\byear{2017})
\end{bchapter}
\endbibitem

\bibitem{inpaint}
\begin{bchapter}
\bauthor{\bsnm{{Pathak}}, \binits{D.}},
\bauthor{\bsnm{{Krähenbühl}}, \binits{P.}},
\bauthor{\bsnm{{Donahue}}, \binits{J.}},
\bauthor{\bsnm{{Darrell}}, \binits{T.}},
\bauthor{\bsnm{{Efros}}, \binits{A.A.}}:
\bctitle{Context encoders: Feature learning by inpainting}.
In: \bbtitle{2016 IEEE Conference on Computer Vision and Pattern Recognition
  (CVPR)},
pp. \bfpage{2536}--\blpage{2544}
(\byear{2016})
\end{bchapter}
\endbibitem

\bibitem{superres}
\begin{bchapter}
\bauthor{\bsnm{{Ledig}}, \binits{C.}},
\bauthor{\bsnm{{Theis}}, \binits{L.}},
\bauthor{\bsnm{{Huszár}}, \binits{F.}},
\bauthor{\bsnm{{Caballero}}, \binits{J.}},
\bauthor{\bsnm{{Cunningham}}, \binits{A.}},
\bauthor{\bsnm{{Acosta}}, \binits{A.}},
\bauthor{\bsnm{{Aitken}}, \binits{A.}},
\bauthor{\bsnm{{Tejani}}, \binits{A.}},
\bauthor{\bsnm{{Totz}}, \binits{J.}},
\bauthor{\bsnm{{Wang}}, \binits{Z.}},
\bauthor{\bsnm{{Shi}}, \binits{W.}}:
\bctitle{Photo-realistic single image super-resolution using a generative
  adversarial network}.
In: \bbtitle{2017 IEEE Conference on Computer Vision and Pattern Recognition
  (CVPR)},
pp. \bfpage{105}--\blpage{114}
(\byear{2017})
\end{bchapter}
\endbibitem

\bibitem{superres2}
\begin{bchapter}
\bauthor{\bsnm{Sønderby}, \binits{C.}},
\bauthor{\bsnm{Caballero}, \binits{J.}},
\bauthor{\bsnm{Theis}, \binits{L.}},
\bauthor{\bsnm{Shi}, \binits{W.}},
\bauthor{\bsnm{Huszár}, \binits{F.}}:
\bctitle{Amortised map inference for image super-resolution}.
In: \bbtitle{International Conference on Learning Representations}
(\byear{2017})
\end{bchapter}
\endbibitem

\bibitem{future}
\begin{bchapter}
\bauthor{\bsnm{Zhou}, \binits{Y.}},
\bauthor{\bsnm{Berg}, \binits{T.}}:
\bctitle{Learning temporal transformations from time-lapse videos},
vol. \bseriesno{9912},
pp. \bfpage{262}--\blpage{277}
(\byear{2016})
\end{bchapter}
\endbibitem

\bibitem{cyclegan}
\begin{bchapter}
\bauthor{\bsnm{Zhu}, \binits{J.-Y.}},
\bauthor{\bsnm{Park}, \binits{T.}},
\bauthor{\bsnm{Isola}, \binits{P.}},
\bauthor{\bsnm{Efros}, \binits{A.A.}}:
\bctitle{Unpaired image-to-image translation using cycle-consistent adversarial
  networks}.
In: \bbtitle{Computer Vision (ICCV), 2017 IEEE International Conference On},
pp. \bfpage{2242}--\blpage{2251}
(\byear{2017})
\end{bchapter}
\endbibitem

\bibitem{dagan}
\begin{bchapter}
\bauthor{\bsnm{{Ma}}, \binits{S.}},
\bauthor{\bsnm{{Fu}}, \binits{J.}},
\bauthor{\bsnm{{Chen}}, \binits{C.W.}},
\bauthor{\bsnm{{Mei}}, \binits{T.}}:
\bctitle{Da-gan: Instance-level image translation by deep attention generative
  adversarial networks}.
In: \bbtitle{2018 IEEE/CVF Conference on Computer Vision and Pattern
  Recognition},
pp. \bfpage{5657}--\blpage{5666}
(\byear{2018})
\end{bchapter}
\endbibitem

\bibitem{photoedit}
\begin{bchapter}
\bauthor{\bsnm{Brock}, \binits{A.}},
\bauthor{\bsnm{Lim}, \binits{T.}},
\bauthor{\bsnm{Ritchie}, \binits{J.}},
\bauthor{\bsnm{Weston}, \binits{N.}}:
\bctitle{Neural Photo Editing With Introspective Adversarial Networks}.
In: \bbtitle{5th International Conference on Learning Representations 2017,
  ICLR 2017},
pp. \bfpage{1}--\blpage{15}
(\byear{2017})
\end{bchapter}
\endbibitem

\bibitem{photoedit2}
\begin{bchapter}
\bauthor{\bsnm{{He}}, \binits{K.}},
\bauthor{\bsnm{{Zhang}}, \binits{X.}},
\bauthor{\bsnm{{Ren}}, \binits{S.}},
\bauthor{\bsnm{{Sun}}, \binits{J.}}:
\bctitle{Deep residual learning for image recognition}.
In: \bbtitle{2016 IEEE Conference on Computer Vision and Pattern Recognition
  (CVPR)},
pp. \bfpage{770}--\blpage{778}
(\byear{2016})
\end{bchapter}
\endbibitem

\bibitem{gan_survey}
\begin{botherref}
\oauthor{\bsnm{Wang}, \binits{Z.}},
\oauthor{\bsnm{She}, \binits{Q.}},
\oauthor{\bsnm{Ward}, \binits{T.E.}}:
Generative adversarial networks in computer vision: A survey and taxonomy.
ACM Comput. Surv.
\textbf{54}(2)
(2021)
\end{botherref}
\endbibitem

\bibitem{img2img_survey}
\begin{barticle}
\bauthor{\bsnm{Chen}, \binits{Y.}},
\bauthor{\bsnm{Zhao}, \binits{Y.}},
\bauthor{\bsnm{Jia}, \binits{W.}},
\bauthor{\bsnm{Cao}, \binits{L.}},
\bauthor{\bsnm{Liu}, \binits{X.}}:
\batitle{Adversarial-learning-based image-to-image transformation: A survey}.
\bjtitle{Neurocomputing}
\bvolume{411},
\bfpage{468}--\blpage{486}
(\byear{2020})
\end{barticle}
\endbibitem

\bibitem{img_synth_and_editing_survey}
\begin{barticle}
\bauthor{\bsnm{Wu}, \binits{X.}},
\bauthor{\bsnm{Xu}, \binits{K.}},
\bauthor{\bsnm{Hall}, \binits{P.}}:
\batitle{A survey of image synthesis and editing with generative adversarial
  networks}.
\bjtitle{Tsinghua Science and Technology}
\bvolume{22}(\bissue{6}),
\bfpage{660}--\blpage{674}
(\byear{2017})
\end{barticle}
\endbibitem

\bibitem{cogan}
\begin{bchapter}
\bauthor{\bsnm{Liu}, \binits{M.-Y.}},
\bauthor{\bsnm{Tuzel}, \binits{O.}}:
\bctitle{Coupled generative adversarial networks}.
In: \beditor{\bsnm{Lee}, \binits{D.}},
\beditor{\bsnm{Sugiyama}, \binits{M.}},
\beditor{\bsnm{Luxburg}, \binits{U.}},
\beditor{\bsnm{Guyon}, \binits{I.}},
\beditor{\bsnm{Garnett}, \binits{R.}} (eds.)
\bbtitle{Advances in Neural Information Processing Systems},
vol. \bseriesno{29}
(\byear{2016})
\end{bchapter}
\endbibitem

\bibitem{stargan}
\begin{bchapter}
\bauthor{\bsnm{{Choi}}, \binits{Y.}},
\bauthor{\bsnm{{Choi}}, \binits{M.}},
\bauthor{\bsnm{{Kim}}, \binits{M.}},
\bauthor{\bsnm{{Ha}}, \binits{J.}},
\bauthor{\bsnm{{Kim}}, \binits{S.}},
\bauthor{\bsnm{{Choo}}, \binits{J.}}:
\bctitle{Stargan: Unified generative adversarial networks for multi-domain
  image-to-image translation}.
In: \bbtitle{2018 IEEE/CVF Conference on Computer Vision and Pattern
  Recognition},
pp. \bfpage{8789}--\blpage{8797}
(\byear{2018})
\end{bchapter}
\endbibitem

\bibitem{munit}
\begin{bchapter}
\bauthor{\bsnm{Huang}, \binits{X.}},
\bauthor{\bsnm{Liu}, \binits{M.-Y.}},
\bauthor{\bsnm{Belongie}, \binits{S.}},
\bauthor{\bsnm{Kautz}, \binits{J.}}:
\bctitle{Multimodal unsupervised image-to-image translation}.
In: \beditor{\bsnm{Ferrari}, \binits{V.}},
\beditor{\bsnm{Hebert}, \binits{M.}},
\beditor{\bsnm{Sminchisescu}, \binits{C.}},
\beditor{\bsnm{Weiss}, \binits{Y.}} (eds.)
\bbtitle{Computer Vision -- ECCV 2018},
pp. \bfpage{179}--\blpage{196}.
\bpublisher{Springer},
\blocation{Cham}
(\byear{2018})
\end{bchapter}
\endbibitem

\bibitem{stargan2}
\begin{bchapter}
\bauthor{\bsnm{Choi}, \binits{Y.}},
\bauthor{\bsnm{Uh}, \binits{Y.}},
\bauthor{\bsnm{Yoo}, \binits{J.}},
\bauthor{\bsnm{Ha}, \binits{J.-W.}}:
\bctitle{Stargan v2: Diverse image synthesis for multiple domains}.
In: \bbtitle{Proceedings of the IEEE/CVF Conference on Computer Vision and
  Pattern Recognition (CVPR)},
pp. \bfpage{8188}--\blpage{8197}
(\byear{2020})
\end{bchapter}
\endbibitem

\bibitem{img2imgsa}
\begin{bchapter}
\bauthor{\bsnm{Kang}, \binits{T.}},
\bauthor{\bsnm{Lee}, \binits{K.}}:
\bctitle{Unsupervised image-to-image translation with self-attention networks}.
In: \bbtitle{2020 IEEE International Conference on Big Data and Smart Computing
  (BigComp)}
(\byear{2020})
\end{bchapter}
\endbibitem

\bibitem{sagan}
\begin{bchapter}
\bauthor{\bsnm{Zhang}, \binits{H.}},
\bauthor{\bsnm{Goodfellow}, \binits{I.}},
\bauthor{\bsnm{Metaxas}, \binits{D.}},
\bauthor{\bsnm{Odena}, \binits{A.}}:
\bctitle{Self-attention generative adversarial networks}.
In: \bbtitle{Proceedings of the 36th International Conference on Machine
  Learning}.
\bsertitle{Proceedings of Machine Learning Research},
pp. \bfpage{7354}--\blpage{7363}
(\byear{2019})
\end{bchapter}
\endbibitem

\bibitem{unet}
\begin{bchapter}
\bauthor{\bsnm{Ronneberger}, \binits{O.}},
\bauthor{\bsnm{Fischer}, \binits{P.}},
\bauthor{\bsnm{Brox}, \binits{T.}}:
\bctitle{U-net: Convolutional networks for biomedical image segmentation}.
In: \beditor{\bsnm{Navab}, \binits{N.}},
\beditor{\bsnm{Hornegger}, \binits{J.}},
\beditor{\bsnm{Wells}, \binits{W.M.}},
\beditor{\bsnm{Frangi}, \binits{A.F.}} (eds.)
\bbtitle{Medical Image Computing and Computer-Assisted Intervention -- MICCAI
  2015},
pp. \bfpage{234}--\blpage{241}.
\bpublisher{Springer},
\blocation{Cham}
(\byear{2015})
\end{bchapter}
\endbibitem

\bibitem{neustyle}
\begin{bchapter}
\bauthor{\bsnm{Johnson}, \binits{J.}},
\bauthor{\bsnm{Alahi}, \binits{A.}},
\bauthor{\bsnm{Fei-Fei}, \binits{L.}}:
\bctitle{Perceptual losses for real-time style transfer and super-resolution}.
In: \bbtitle{European Conference on Computer Vision}
(\byear{2016})
\end{bchapter}
\endbibitem

\bibitem{lenet}
\begin{bchapter}
\bauthor{\bsnm{Lecun}, \binits{Y.}},
\bauthor{\bsnm{Bottou}, \binits{L.}},
\bauthor{\bsnm{Bengio}, \binits{Y.}},
\bauthor{\bsnm{Haffner}, \binits{P.}}:
\bctitle{Gradient-based learning applied to document recognition}.
In: \bbtitle{Proceedings of the IEEE},
pp. \bfpage{2278}--\blpage{2324}
(\byear{1998})
\end{bchapter}
\endbibitem

\bibitem{wganobj}
\begin{bchapter}
\bauthor{\bsnm{Gulrajani}, \binits{I.}},
\bauthor{\bsnm{Ahmed}, \binits{F.}},
\bauthor{\bsnm{Arjovsky}, \binits{M.}},
\bauthor{\bsnm{Dumoulin}, \binits{V.}},
\bauthor{\bsnm{Courville}, \binits{A.}}:
\bctitle{Improved training of wasserstein gans}.
In: \bbtitle{Proceedings of the 31st International Conference on Neural
  Information Processing Systems}.
\bsertitle{NIPS’17},
pp. \bfpage{5769}--\blpage{5779}.
\bpublisher{Curran Associates Inc.},
\blocation{Red Hook, NY, USA}
(\byear{2017})
\end{bchapter}
\endbibitem

\bibitem{unit}
\begin{bchapter}
\bauthor{\bsnm{Liu}, \binits{M.-Y.}},
\bauthor{\bsnm{Breuel}, \binits{T.}},
\bauthor{\bsnm{Kautz}, \binits{J.}}:
\bctitle{Unsupervised image-to-image translation networks}.
In: \beditor{\bsnm{Guyon}, \binits{I.}},
\beditor{\bsnm{Luxburg}, \binits{U.V.}},
\beditor{\bsnm{Bengio}, \binits{S.}},
\beditor{\bsnm{Wallach}, \binits{H.}},
\beditor{\bsnm{Fergus}, \binits{R.}},
\beditor{\bsnm{Vishwanathan}, \binits{S.}},
\beditor{\bsnm{Garnett}, \binits{R.}} (eds.)
\bbtitle{Advances in Neural Information Processing Systems},
vol. \bseriesno{30}
(\byear{2017})
\end{bchapter}
\endbibitem

\bibitem{bicyclegan}
\begin{bchapter}
\bauthor{\bsnm{Zhu}, \binits{J.-Y.}},
\bauthor{\bsnm{Zhang}, \binits{R.}},
\bauthor{\bsnm{Pathak}, \binits{D.}},
\bauthor{\bsnm{Darrell}, \binits{T.}},
\bauthor{\bsnm{Efros}, \binits{A.A.}},
\bauthor{\bsnm{Wang}, \binits{O.}},
\bauthor{\bsnm{Shechtman}, \binits{E.}}:
\bctitle{Toward multimodal image-to-image translation}.
In: \bbtitle{NIPS}
(\byear{2017})
\end{bchapter}
\endbibitem

\bibitem{fid}
\begin{bchapter}
\bauthor{\bsnm{Heusel}, \binits{M.}},
\bauthor{\bsnm{Ramsauer}, \binits{H.}},
\bauthor{\bsnm{Unterthiner}, \binits{T.}},
\bauthor{\bsnm{Nessler}, \binits{B.}},
\bauthor{\bsnm{Hochreiter}, \binits{S.}}:
\bctitle{Gans trained by a two time-scale update rule converge to a local nash
  equilibrium}.
In: \bbtitle{Proceedings of the 31st International Conference on Neural
  Information Processing Systems}.
\bsertitle{NIPS’17},
pp. \bfpage{6629}--\blpage{6640}.
\bpublisher{Curran Associates Inc.},
\blocation{Red Hook, NY, USA}
(\byear{2017})
\end{bchapter}
\endbibitem

\bibitem{precrec}
\begin{bchapter}
\bauthor{\bsnm{Sajjadi}, \binits{M.S.M.}},
\bauthor{\bsnm{Bachem}, \binits{O.}},
\bauthor{\bsnm{Lucic}, \binits{M.}},
\bauthor{\bsnm{Bousquet}, \binits{O.}},
\bauthor{\bsnm{Gelly}, \binits{S.}}:
\bctitle{Assessing generative models via precision and recall}.
In: \bbtitle{Proceedings of the 32nd International Conference on Neural
  Information Processing Systems}.
\bsertitle{NIPS’18},
pp. \bfpage{5234}--\blpage{5243}.
\bpublisher{Curran Associates Inc.},
\blocation{Red Hook, NY, USA}
(\byear{2018})
\end{bchapter}
\endbibitem

\bibitem{mnist}
\begin{barticle}
\bauthor{\bsnm{Deng}, \binits{L.}}:
\batitle{The mnist database of handwritten digit images for machine learning
  research}.
\bjtitle{IEEE Signal Processing Magazine}
\bvolume{29}(\bissue{6}),
\bfpage{141}--\blpage{142}
(\byear{2012})
\end{barticle}
\endbibitem

\bibitem{usps}
\begin{bchapter}
\bauthor{\bsnm{Carlucci}, \binits{F.}},
\bauthor{\bsnm{Russo}, \binits{P.}},
\bauthor{\bsnm{Tommasi}, \binits{T.}},
\bauthor{\bsnm{Caputo}, \binits{B.}}:
\bctitle{Hallucinating agnostic images to generalize across domains}.
In: \bbtitle{2019 IEEE/CVF International Conference on Computer Vision Workshop
  (ICCVW)},
pp. \bfpage{3227}--\blpage{3234}.
\bpublisher{IEEE Computer Society},
\blocation{Los Alamitos, CA, USA}
(\byear{2019})
\end{bchapter}
\endbibitem

\bibitem{proscons}
\begin{barticle}
\bauthor{\bsnm{Borji}, \binits{A.}}:
\batitle{Pros and cons of gan evaluation measures}.
\bjtitle{Computer Vision and Image Understanding}
\bvolume{179},
\bfpage{41}--\blpage{65}
(\byear{2019})
\end{barticle}
\endbibitem

\bibitem{noteincep}
\begin{botherref}
\oauthor{\bsnm{Barratt}, \binits{S.T.}},
\oauthor{\bsnm{Sharma}, \binits{R.}}:
A note on the inception score.
ArXiv
\textbf{abs/1801.01973}
(2018)
\end{botherref}
\endbibitem

\bibitem{tragan}
\begin{bchapter}
\bauthor{\bsnm{Salimans}, \binits{T.}},
\bauthor{\bsnm{Goodfellow}, \binits{I.}},
\bauthor{\bsnm{Zaremba}, \binits{W.}},
\bauthor{\bsnm{Cheung}, \binits{V.}},
\bauthor{\bsnm{Radford}, \binits{A.}},
\bauthor{\bsnm{Chen}, \binits{X.}},
\bauthor{\bsnm{Chen}, \binits{X.}}:
\bctitle{Improved techniques for training gans}.
In: \beditor{\bsnm{Lee}, \binits{D.D.}},
\beditor{\bsnm{Sugiyama}, \binits{M.}},
\beditor{\bsnm{Luxburg}, \binits{U.V.}},
\beditor{\bsnm{Guyon}, \binits{I.}},
\beditor{\bsnm{Garnett}, \binits{R.}} (eds.)
\bbtitle{Advances in Neural Information Processing Systems 29},
pp. \bfpage{2234}--\blpage{2242}
(\byear{2016})
\end{bchapter}
\endbibitem

\bibitem{imnettext}
\begin{bchapter}
\bauthor{\bsnm{Geirhos}, \binits{R.}},
\bauthor{\bsnm{Rubisch}, \binits{P.}},
\bauthor{\bsnm{Michaelis}, \binits{C.}},
\bauthor{\bsnm{Bethge}, \binits{M.}},
\bauthor{\bsnm{Wichmann}, \binits{F.A.}},
\bauthor{\bsnm{Brendel}, \binits{W.}}:
\bctitle{Imagenet-trained {CNN}s are biased towards texture; increasing shape
  bias improves accuracy and robustness.}
In: \bbtitle{International Conference on Learning Representations}
(\byear{2019})
\end{bchapter}
\endbibitem

\bibitem{percepmetr}
\begin{bchapter}
\bauthor{\bsnm{{Zhang}}, \binits{R.}},
\bauthor{\bsnm{{Isola}}, \binits{P.}},
\bauthor{\bsnm{{Efros}}, \binits{A.A.}},
\bauthor{\bsnm{{Shechtman}}, \binits{E.}},
\bauthor{\bsnm{{Wang}}, \binits{O.}}:
\bctitle{The unreasonable effectiveness of deep features as a perceptual
  metric}.
In: \bbtitle{2018 IEEE/CVF Conference on Computer Vision and Pattern
  Recognition},
pp. \bfpage{586}--\blpage{595}
(\byear{2018})
\end{bchapter}
\endbibitem

\bibitem{cityscapes}
\begin{bchapter}
\bauthor{\bsnm{{Cordts}}, \binits{M.}},
\bauthor{\bsnm{{Omran}}, \binits{M.}},
\bauthor{\bsnm{{Ramos}}, \binits{S.}},
\bauthor{\bsnm{{Rehfeld}}, \binits{T.}},
\bauthor{\bsnm{{Enzweiler}}, \binits{M.}},
\bauthor{\bsnm{{Benenson}}, \binits{R.}},
\bauthor{\bsnm{{Franke}}, \binits{U.}},
\bauthor{\bsnm{{Roth}}, \binits{S.}},
\bauthor{\bsnm{{Schiele}}, \binits{B.}}:
\bctitle{The cityscapes dataset for semantic urban scene understanding}.
In: \bbtitle{2016 IEEE Conference on Computer Vision and Pattern Recognition
  (CVPR)},
pp. \bfpage{3213}--\blpage{3223}
(\byear{2016})
\end{bchapter}
\endbibitem

\bibitem{facades}
\begin{bchapter}
\bauthor{\bsnm{Tyle{\v c}ek}, \binits{R.}},
\bauthor{\bsnm{{\v S}{\' a}ra}, \binits{R.}}:
\bctitle{Spatial pattern templates for recognition of objects with regular
  structure}.
In: \bbtitle{Proc. GCPR},
\bconflocation{Saarbrucken, Germany}
(\byear{2013})
\end{bchapter}
\endbibitem

\bibitem{imgnet}
\begin{bchapter}
\bauthor{\bsnm{{Deng}}, \binits{J.}},
\bauthor{\bsnm{{Dong}}, \binits{W.}},
\bauthor{\bsnm{{Socher}}, \binits{R.}},
\bauthor{\bsnm{{Li}}, \binits{L.}},
\bauthor{\bsnm{{Kai Li}}},
\bauthor{\bsnm{{Li Fei-Fei}}}:
\bctitle{Imagenet: A large-scale hierarchical image database}.
In: \bbtitle{2009 IEEE Conference on Computer Vision and Pattern Recognition},
pp. \bfpage{248}--\blpage{255}
(\byear{2009})
\end{bchapter}
\endbibitem

\bibitem{wordnet}
\begin{barticle}
\bauthor{\bsnm{Miller}, \binits{G.A.}}:
\batitle{Wordnet: A lexical database for english}.
\bjtitle{COMMUNICATIONS OF THE ACM}
\bvolume{38},
\bfpage{39}--\blpage{41}
(\byear{1995})
\end{barticle}
\endbibitem

\bibitem{utzappos}
\begin{bchapter}
\bauthor{\bsnm{Yu}, \binits{A.}},
\bauthor{\bsnm{Grauman}, \binits{K.}}:
\bctitle{Fine-grained visual comparisons with local learning}.
In: \bbtitle{Computer Vision and Pattern Recognition (CVPR)},
pp. \bfpage{192}--\blpage{199}
(\byear{2014})
\end{bchapter}
\endbibitem

\bibitem{transient}
\begin{botherref}
\oauthor{\bsnm{Laffont}, \binits{P.-Y.}},
\oauthor{\bsnm{Ren}, \binits{Z.}},
\oauthor{\bsnm{Tao}, \binits{X.}},
\oauthor{\bsnm{Qian}, \binits{C.}},
\oauthor{\bsnm{Hays}, \binits{J.}}:
Transient attributes for high-level understanding and editing of outdoor
  scenes.
ACM Transactions on Graphics (proceedings of SIGGRAPH)
\textbf{33}(4)
(2014)
\end{botherref}
\endbibitem

\bibitem{sketch}
\begin{barticle}
\bauthor{\bsnm{Eitz}, \binits{M.}},
\bauthor{\bsnm{Hays}, \binits{J.}},
\bauthor{\bsnm{Alexa}, \binits{M.}}:
\batitle{How do humans sketch objects?}
\bjtitle{ACM Trans. Graph. (Proc. SIGGRAPH)}
\bvolume{31}(\bissue{4}),
\bfpage{44}--\blpage{14410}
(\byear{2012})
\end{barticle}
\endbibitem

\bibitem{multispec}
\begin{bchapter}
\bauthor{\bsnm{{Hwang}}, \binits{S.}},
\bauthor{\bsnm{{Park}}, \binits{J.}},
\bauthor{\bsnm{{Kim}}, \binits{N.}},
\bauthor{\bsnm{{Choi}}, \binits{Y.}},
\bauthor{\bsnm{{Kweon}}, \binits{I.S.}}:
\bctitle{Multispectral pedestrian detection: Benchmark dataset and baseline}.
In: \bbtitle{2015 IEEE Conference on Computer Vision and Pattern Recognition
  (CVPR)},
pp. \bfpage{1037}--\blpage{1045}
(\byear{2015})
\end{bchapter}
\endbibitem

\bibitem{celeba}
\begin{bchapter}
\bauthor{\bsnm{{Liu}}, \binits{Z.}},
\bauthor{\bsnm{{Luo}}, \binits{P.}},
\bauthor{\bsnm{{Wang}}, \binits{X.}},
\bauthor{\bsnm{{Tang}}, \binits{X.}}:
\bctitle{Deep learning face attributes in the wild}.
In: \bbtitle{2015 IEEE International Conference on Computer Vision (ICCV)},
pp. \bfpage{3730}--\blpage{3738}
(\byear{2015})
\end{bchapter}
\endbibitem

\bibitem{nyu}
\begin{bchapter}
\bauthor{\bsnm{Nathan~Silberman}, \binits{P.K.} \bsuffix{Derek~Hoiem}},
\bauthor{\bsnm{Fergus}, \binits{R.}}:
\bctitle{Indoor segmentation and support inference from rgbd images}.
In: \bbtitle{ECCV}
(\byear{2012})
\end{bchapter}
\endbibitem

\bibitem{rgbd}
\begin{bchapter}
\bauthor{\bsnm{Lai}, \binits{K.}},
\bauthor{\bsnm{Bo}, \binits{L.}},
\bauthor{\bsnm{Ren}, \binits{X.}},
\bauthor{\bsnm{Fox}, \binits{D.}}:
\bctitle{A large-scale hierarchical multi-view rgb-d object dataset}.
In: \bbtitle{2011 IEEE International Conference on Robotics and Automation},
pp. \bfpage{1817}--\blpage{1824}
(\byear{2011})
\end{bchapter}
\endbibitem

\bibitem{synthia}
\begin{bchapter}
\bauthor{\bsnm{Ros}, \binits{G.}},
\bauthor{\bsnm{Sellart}, \binits{L.}},
\bauthor{\bsnm{Materzynska}, \binits{J.}},
\bauthor{\bsnm{Vazquez}, \binits{D.}},
\bauthor{\bsnm{Lopez}, \binits{A.M.}}:
\bctitle{The synthia dataset: A large collection of synthetic images for
  semantic segmentation of urban scenes}.
In: \bbtitle{2016 IEEE Conference on Computer Vision and Pattern Recognition
  (CVPR)},
pp. \bfpage{3234}--\blpage{3243}
(\byear{2016})
\end{bchapter}
\endbibitem

\bibitem{rafd}
\begin{barticle}
\bauthor{\bsnm{Langner}, \binits{O.}},
\bauthor{\bsnm{Dotsch}, \binits{R.}},
\bauthor{\bsnm{Bijlstra}, \binits{G.}},
\bauthor{\bsnm{Wigboldus}, \binits{D.}},
\bauthor{\bsnm{Hawk}, \binits{S.}},
\bauthor{\bsnm{Knippenberg}, \binits{A.}}:
\batitle{Presentation and validation of the radboud face database}.
\bjtitle{Cognition \& Emotion - COGNITION EMOTION}
\bvolume{24},
\bfpage{1377}--\blpage{1388}
(\byear{2010})
\end{barticle}
\endbibitem

\bibitem{proggan}
\begin{bchapter}
\bauthor{\bsnm{Karras}, \binits{T.}},
\bauthor{\bsnm{Aila}, \binits{T.}},
\bauthor{\bsnm{Laine}, \binits{S.}},
\bauthor{\bsnm{Lehtinen}, \binits{J.}}:
\bctitle{Progressive growing of {GAN}s for improved quality, stability, and
  variation}.
In: \bbtitle{International Conference on Learning Representations}
(\byear{2018})
\end{bchapter}
\endbibitem

\bibitem{facescrub}
\begin{bchapter}
\bauthor{\bsnm{{Ng}}, \binits{H.}},
\bauthor{\bsnm{{Winkler}}, \binits{S.}}:
\bctitle{A data-driven approach to cleaning large face datasets}.
In: \bbtitle{2014 IEEE International Conference on Image Processing (ICIP)},
pp. \bfpage{343}--\blpage{347}
(\byear{2014})
\end{bchapter}
\endbibitem

\bibitem{cub200}
\begin{botherref}
\oauthor{\bsnm{Wah}, \binits{C.}},
\oauthor{\bsnm{Branson}, \binits{S.}},
\oauthor{\bsnm{Welinder}, \binits{P.}},
\oauthor{\bsnm{Perona}, \binits{P.}},
\oauthor{\bsnm{Belongie}, \binits{S.}}:
{The Caltech-UCSD Birds-200-2011 Dataset}.
Technical Report CNS-TR-2011-001,
California Institute of Technology
(2011)
\end{botherref}
\endbibitem

\end{thebibliography}

\end{document}